# Unsupervised Frequent Pattern Mining for CEP

Guy Shapira

# Unsupervised Frequent Pattern Mining for CEP

Research Thesis

Submitted in partial fulfillment of the requirements
for the degree of Master of Science in Computer Science

## Guy Shapira

The Technion's funding of this research is hereby acknowledged.

# Contents

# List of Figures

# Abstract

Complex Event Processing (CEP) is a set of methods that allow efficient knowledge extraction from massive data streams using complex and highly descriptive patterns. Numerous applications, such as online finance, healthcare monitoring and fraud detection use CEP technologies to capture critical alerts, potential threats, or vital notifications in real time. As of today, in many fields, *patterns* are manually defined by human experts. However, desired patterns often contain convoluted relations that are difficult for humans to detect, and human expertise is scarce in many domains.

We present REDEEMER (<u>RE</u>inforcement base<u>D</u> c<u>E</u>p patt<u>E</u>rn <u>M</u>in<u>ER</u>), a novel reinforcement and active learning approach aimed at mining CEP patterns that allow expansion of the knowledge extracted while reducing the human effort required. This approach includes a novel policy gradient method for vast multivariate spaces and a new way to combine reinforcement and active learning for CEP rule learning while minimizing the number of labels needed for training.

REDEEMER aims to enable CEP integration in domains that could not utilize it before. To the best of our knowledge, REDEEMER is the first system that suggests new CEP rules that were not observed beforehand, and is the first method aimed for increasing pattern knowledge in fields where experts do not possess sufficient information required for CEP tools.

Our experiments on diverse data-sets demonstrate that REDEEMER is able to extend pattern knowledge while outperforming several *state-of-the-art* reinforcement learning methods for pattern mining.

# Chapter 1

# Introduction

Recent years have seen significant development in high-speed data stream analysis, with methods for efficient analysis being applied in areas such as online finance [ABNS06], healthcare monitoring [BEE+10], and fraud detection [WvAHS09]. Complex Event Processing (CEP) has emerged as a valuable tool for extracting information from massive data streams by real-time pattern matching. Every CEP pattern is a sequence of related events, meeting some temporal constraints. Patterns describe domain specific key phenomena in the data stream and provide valuable insights into important behavioral characteristics of the data. Therefore, a core feature in every CEP system is a declarative event processing language used to express patterns in data streams [BDO19].

Although the relevance of CEP systems for many domains has been highlighted in previous research, it is difficult and even sometimes impossible to employ them in various scenarios. The burden of defining CEP patterns is time-consuming and requires extensive domain knowledge, as a consequence of which integration of CEP systems is hampered in many fields.

Prior studies [BDO19, PVS11, MKE+15] noted the disadvantages of manually defined CEP rules. However, the solutions suggested so far rely on a domain expert locating key phenomena shortly after they occur and marking these findings for a learning model [BDO19, MTZ17], or they assume the expert can distinguish between key phenomena in real-time and label them accordingly [MKE+15, YK09]. In both cases, the proposed solution reduces the complexity of the expert work but still requires an extensive and sometimes unattainable understanding of the relationships between events. More importantly, because of those assumptions, all existing solutions focus on the problem of discovery and refinement of known patterns that had already been observed, while we are interested in the generation of new meaningful patterns that no prior knowledge of them is available.

In rare cases where an expert is available and possesses sufficient prior knowledge, the dependency on human experts is still problematic since it lowers the system's reliability and limits the complexity of the patterns that can be found. As the possible set of patterns is almost limitless, and data streams are often multidimensional and sophisticated, the best patterns can be too complex for humans to define. Consequently, there is a compelling need to bridge the gap between CEP and machine learning approaches that automatically define rules.

In this work, we propose a novel method that allows pattern mining in complex, high-speed data streams while requiring minimal domain knowledge. We present *RE-DEEMER* (<u>RE</u>inforcement base<u>D</u> c<u>E</u>p patt<u>E</u>rn <u>Min</u><u>ER</u>), an innovative method based on reinforcement and active learning that automatically derives meaningful and frequent CEP rules that were not previously observed or known.

**Contributions**. Our main contributions can be summarized as follows:

- We formally define the problem of pattern knowledge expansion when prior knowledge is lacking, and expert availability is limited.

- We propose a new policy gradient method that is aimed for split-structured *Actor-Critic* reinforcement learning models.

- We develop REDEEMER, a semi-supervised learning model that generates not previously known meaningful patterns while also minimizing expert interaction.

- We present experiments on diverse data sets to confirm that REDEEMER outperforms existing methods that were adapted for the new problem as well as to demonstrate the practicality of REDEEMER for realistic use-cases.

**Outline.** The remainder of the thesis is structured as follows. Chapter 2 surveys related work on pattern mining, CEP in general, and automatic rule-learning for CEP in particular. Chapter 3 provides background on CEP, reinforcement, and active learning methods. In Chapter 4 formally presents our problem setting, defines notations used throughout our work, and explains the key metrics used to evaluate this work. In Chapter 5, we introduce REDEEMER, the first automatic rule generation system to create new CEP rules based on known patterns of interest. In Chapter 6, we conduct an extensive experimental evaluation of REDEEMER, compare it to existing reinforcement learning solutions as well as evaluate its performance in real use cases. Chapter 7 presents our conclusions.

# Chapter 2

# Related Work

**CEP.** In recent years, there has been an increasing interest in adapting pattern mining approaches to automatically learn **CEP rules** without intervention from domain experts. The autoCEP system [MTZ17] is a data mining based approach that automatically derives CEP patterns from prerecorded labeled traces. It exploits shapelets learning [YK09] instead of learning complete rules. Another recently presented framework is iCEP [MKE+15], which splits rule generation into multiple sub-modules, each responsible for learning different parts of the pattern. The authors of [BDO19] introduced GP4CEP, the first solution to employ genetic programming to generate CEP patterns. GP4CEP learns rules end-to-end using evolutionary operations (usually mutation techniques and selection by fitness) to deliver state-of-the-art results, even when used on a vast search space.

Our work differs from previous work regarding the role of the expert in the mining process and in the overall task learned. For example, GP4CEP uses rule inference from annotated points of interest in prerecorded data and yields a pattern that occurs in all points of interest, while we aim to expand pattern knowledge by generating new CEP rules that are not known and therefore could not have been annotated beforehand. Nonetheless, several concepts from GP4CEP strongly influenced our approach, namely their solutions to pattern representation issues and the design of the syntax tree.

In summary, the complexity of learning CEP rules remains an open research problem [BDO19, MKE+15]. Moreover, using reinforcement and active learning methods for automatic CEP rule generation is a new and innovative approach that enables new directions that are aimed at mining patterns that were not priorly observed.

**Pattern Mining.** Pattern mining is a broad term that refers to techniques for automated knowledge extraction from large databases or data-streams. Among others, important techniques in this area include item-set mining [FVLV+17], which aims to find items that often occur together, and frequent sequential pattern mining [FVLR+17], which aims to find items that frequently occur in a specific fixed order. Early algorithms in the field of pattern mining include the one introduced in [AIS93], which is used to mine rules that include frequent associations between sets of items, as well as FP-Growth [HP00] and ECLAT [Zak00], which are efficient alternatives to frequent rule mining. Further studies presented new solutions that support multidimensional data [GA19, LQSC12, YC05].

As modern applications grow dramatically more sophisticated and operate on highly multidimensional data, the demand for efficient and expressive pattern mining increases. However, current state-of-the-art solutions are not applicable for such complex data streams. Most current frequent pattern mining methods cannot be efficiently adapted for big and complex data streams [AH14]. Because of those limitations, we could not have used similar methods to traverse the pattern space and relied upon several learning techniques to find efficient alternatives.

**Active Learning.** Active Learning (AL) is an iterative supervised machine learning training scheme used with datasets that have little or no ground truth labels. Labels for supervised training are created by a label source (a human or an algorithm), sometimes called *oracle* or *teacher*. On each training iteration, new unlabeled data points are chosen, then the oracle is interactively queried for labels for the new data points, and the labels are used in a supervised learning training scheme to improve the learning model's performance [SZGW20].

Often AL is tasked with constructing a small training set that represents the whole example space and can be used to achieve satisfactory results [ZCMR21]. In [RXC+21], the significance of AL to deep learning scenarios of high uncertainty is presented. The authors highlight use cases in which the oracle's prediction confidence was used to determine the query strategy. Similar findings were presented in [DP18], who conducted extensive experiments over state-of-the-art AL techniques in multiple scenarios and found uncertainty based methods to be competitive in all of them.

In this work, we deal with the scenario where the domain expert does not have sufficient prior knowledge about the full pattern space; therefore, uncertainty based strategies were utilized to gather knowledge about it.

# Chapter 3

# Preliminaries

## 3.1 CEP

CEP systems observe real-time data streams for combinations of events and conditions that satisfy domain-specific patterns. Such patterns may include a combination of primitive events, operators, and predicates occurring within a predetermined time window [KS18a]. Every time window consists of a fixed number of consecutive records from the data-set.

**Pattern Structure.** Each pattern contains multiple events; each event has a unique identifier and a set of conditions on its attributes. For example, in the output rule appearing in Figure 4.1, the pattern has five events, and there are also conditions over some of those events' attributes.

Patterns are constructed over events using an SQL-like language, where the predicates to be satisfied by the participating events are usually organized in a Boolean formula [KS18b]. Any condition can be defined on any of the attributes of a single event, between an attribute and a numeric value, or between attributes of two distinct events. The most commonly used operators for CEP patterns are the sequence operator (SEQ), which defines both the conditions on its events and their order, the conjunction operator (AND), and the disjunction operator (OR).

We use a pattern structure suggested by the authors of [WDR06]:

    Events    < events pattern >
    Where  < conditions over events' attributes >
    Within    < time window >

As an example, the following pattern can be defined for a CCTV control system for a bank's vault: camera A detects a suspicious person entering its field of view, later camera B located in the adjacent room detects the same person, and finally, camera C located over the exit door detects the person exiting the facility, all within 20 seconds of the first detection by camera A. This pattern is presented by the aforementioned structure in the following:

| Events | SEQ(A as a, B as b, C as c) |
|---|---|
| Where | $((a.person\_id = b.person\_id) \wedge$ |
|  | $(b.person\_id = c.person\_id))$ |
| Within | 20 seconds |

**Usage.** When CEP systems are supplied with domain-specific patterns as input and are connected to a high-speed data stream, the system employs a variety of evaluation mechanisms (e.g., trees [MM09], NFAs [WDR06] and graphs [AcT08]) to find partial matches and combine these to efficiently locate matched patterns in the stream. CEP has already been applied in a number of domains (e.g., healthcare monitoring [BEE+10], fraud detection [WvAHS09], and online finance [ABNS06]) and achieved desirable results when tasked with detection of already well-known patterns.

## 3.2 Reinforcement Learning

The authors of [SB18] defined Reinforcement Learning (RL) as a computational approach that mimics the way humans learn by interacting with their environment. A general reinforcement learning framework usually includes three important components: An *agent* interacts with the environment by performing *actions* according to a deterministic or stochastic policy. An *environment* that is out of the agent's direct control (cannot be altered directly) but is responsive to its actions and provides rewards based on its state. A *reward*, which is a predefined metric that represents how well the agent interacts with the environment.

We summarized all reinforcement learning definitions essential for understanding this work in Table 3.1. We also present all the notations used in our work w.r.t to reinforcement learning in Table 3.2.

**Actor-Critic RL** Actor-Critic (AC) methods use two separate structures to represent a policy that is independent of the value function (estimated reward) [SB18]. The *policy* structure is also known as the *actor* since it selects actions (and thus defines the policy for action selection). The second structure estimates the state value and is also referred to as *critic* because its evaluation criticizes the actor's actions.

Using the standard notation of $(\mathcal{S}, \mathcal{A}, G_t)$ in which $\mathcal{S} \subseteq \mathbb{R}^n$ is the state space (i.e., all possible states the agent might encounter), $\mathcal{A} \subseteq \mathbb{R}^m$ is the action space (i.e., all possible actions that can be employed), and $G_t$ is the expected future reward, we define the policy gradient ($\nabla \mathcal{J}$) w.r.t. $\theta$ as the following:

$$\nabla_\theta \mathcal{J}(\theta) = \mathbb{E}_\pi [\sum_{t=0}^{T-1} \nabla_\theta \log(\pi_\theta(a_t|s_t)) G_t]$$

Where $a_t \in \mathcal{A}, s_t \in \mathcal{S}$ and $\pi_\theta$ is a policy parameterized by $\theta$.

| term | definition | meaning in pattern generation |
|------|------------|------------------------------|
| state | The state describes the current situation, this includes all things that are observed by the agent. | A time window of a data-stream that is received by the control unit of the system. |
| action | The agent's methods which allow it to interact and change its environment, and alter the current visible state | The group of all possible events and conditions that can be added to an existing pattern. |
| policy | A function that returns a feasible action when given the current state. | |
| policy gradient | A method responsible for modeling and optimizing the policy directly. | |
| advantage function | The estimated gain for taking a specific action compared to the average reward expected for all possible actions at the given state. | |
| value-function | Expected reward at a given scenario, represents how good the scenario is for the agent. | Pattern value. Metric that evaluates the pattern relevancy for a domain expert. |
| terminal state | The end of the reinforcement learning episode, there is nothing in the future for the agent to do or to learn from. | End of pattern generation. |

Table 3.1: Reinforcement learning definitions and meaning in pattern generation terms.

**Advantage Actor-Critic.** In *Advantage Actor-Critic* (A2C) [MBM+16], we use the state-value function, notated by $V(s)$ and the action-value function, notated by $Q(s, a)$ as baseline functions. We then define the *advantage value* $A(s_t, a_t)$ using the following formula:

$$A(s_t, a_t) := Q(s_t, a_t) - V(s_t) = r_{t+1} + \gamma V(s_{t+1}) - V(s_t)$$

where $r_{t+1}$ is the reward for the $t + 1$ step and $\gamma$ is the predetermined discount factor. The advantage is compared to the average expected reward of all possible actions in the given state. Using these notations, the authors of [MBM+16] defined the policy gradient ($\mathcal{J}$) w.r.t. $\theta$ as follows:

$$\nabla_\theta \mathcal{J}(\theta) = \mathbb{E}_\pi[\Sigma_{t=0}^{T-1} \nabla_\theta \log(\pi_\theta(a_t|s_t)) A(s_t, a_t)] \tag{3.1}$$

| symbol | meaning |
|--------|---------|
| $\mathcal{S}$ | The complete state space. |
| $\mathcal{A}$ | The complete action space. |
| $s_t$ | The state observed at time stamp $t$. |
| $a_t$ | The action performed at time stamp $t$. |
| $G_t$ | Expected reward at state $s_t$. |
| $\theta$ | The agent's neural network parameters. |
| $\pi$ | The policy of an agent. |
| $\pi_\theta$ | A policy parameterized by a network's parameters $\theta$. |
| $\nabla \mathcal{J}$ | The policy gradient. |
| $\nabla \mathcal{J}(\theta)$ | The policy gradient w.r.t. a network's parameters $\theta$. |
| $V(s)$ | The expected reward (value-function) at state $s$. |
| $Q(s, a)$ | Expected reward (value-function) at state $s$ when action $a$ is performed. |
| $A(s, t)$ | The advantage function when state $s$ is observed and action $a$ is performed. |
| $\gamma$ | Discount factor penalty to uncertainty of future rewards. |

Table 3.2: Symbol table in reinforcement learning used in our work. Adapted from [Wen18].

The advantage function encourages the model to exploit better actions since it weighs the policy's actions according to how advantageous they are compared with other possible actions. At every step, the reward gained by the action the was selected by the agent is compared with all other actions, and the advantage is the difference between those two values (positive if better than the general actions and negative otherwise). In A2C, the actor's parameters are updated according to the policy gradient, and the critic's parameters are updated according to a loss function(e.g., the mean squared error (MSE)).

An important advantage of the AC and A2C methods was presented by the authors of [MSIB20], who showed that they both are applicable for active learning scenarios. Since we aim to incorporate interactions with an expert throughout the learning process, applicability for active learning was crucial.

## 3.3 Active Learning

The main idea behind active learning is to produce accurate models while requiring less labeled data [Set09]. Active learning models can interactively query a human expert in order to receive labels for specific new data points. In most cases, an active learning model will start with a small labeled data set and will incrementally enlarge it by querying an expert. The authors of [WLC+20] have shown that when active learning is used together with proper sampling methods, it reduces the amount of labeled data required for training and achieves better accuracy in cases of imbalanced data.

*Uncertainty-based* query strategies focus on finding examples where the confidence in prediction is the lowest [Set09]. Usually, low-confidence data points belong to classes that are not sufficiently represented in the labeled set; beyond resulting in low confidence for the prediction, this misrepresentation often leads to an inaccurate prediction. When using uncertainty-based strategies, a more diverse training set can be achieved, leading to better performance overall data points.

Since we deal with the scenarios where information about the majority of patterns is nonexistent at initialization, uncertainty-based methods were utilized to gather knowledge about the pattern space.

## 3.4 UCB

The Upper Confidence Bound (known as UCB1) is one of the most commonly used solutions for multi-armed bandit problems. It is based upon the principle of optimism when faced with uncertainty [ACBF02]. Previous research [Aue03] found UCB1 useful when facing the exploitation-exploration trade-off. More formally: Let $N_t(a)$ be the number of times the $a$ action was selected so far, and $s$ be the current state. After $t-1$ actions have been played; we choose the $t^{th}$ action using the following formula:

$$a_t = \arg\max_a (Q(s, a) + c\sqrt{\log(t)/N_t(a)}) \tag{3.2}$$

By this alteration to the vanilla action selection formula, UCB1 reduces the likelihood of actions that were selected many times to be selected again while also encouraging the agent to select actions that were rarely exploited.

# Chapter 4

# Problem Definition

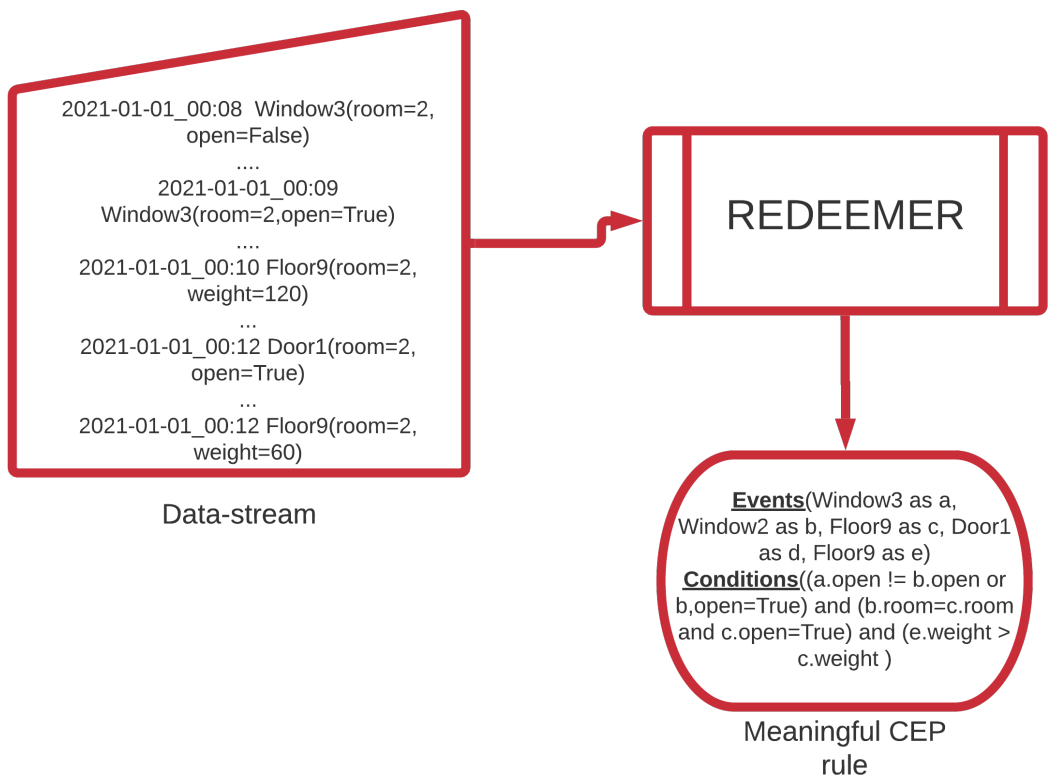

Data-stream

REDEEMER

**Events**(Window3 as a, Window2 as b, Floor9 as c, Door1 as d, Floor9 as e)
**Conditions**((a.open != b.open or b,open=True) and (b.room=c.room and c.open=True) and (e.weight > c.weight )

Meaningful CEP rule

Figure 4.1: Use case example. Raw input tracked from multiple sensors located in a smart home is received by the model; REDEEMER output is parsed as a CEP pattern that describes an intruder entering through the window.

Our main goal is to accurately **expand pattern knowledge** while **minimizing the amount of work needed** by a domain expert. To achieve this, we aim to develop a semi-supervised system that derives CEP rules automatically. In CEP terms, this is represented by the generation of meaningful and frequent patterns that describe interesting situations in the data-stream.

Our approach relies on the existence of a minimal training data-set that contains examples of priorly known patterns along with their numerical rankings; these rankings indicate how interesting each pattern is for a domain expert. Another assumption we rely on is the expert's availability and ability to further rank new patterns that will be suggested by a learning model.

Take for example, a smart home equipped with sensors on the windows, floors, and doors. **Our assumption** would be the availability of some simple patterns, such as a window opened and closed several times within the span of one minute, someone moving between multiple rooms within seconds, etc. We also assume the existence of knowledge about how interesting these patterns are. As part of the learning process, we also require an expert who is familiar with the sensors in the system and is familiar with some simple patterns common in smart homes based systems to be available to answer queries.

**Our goal** would be to find previously unknown frequent and meaningful patterns over the domain. An example of one such pattern that describes an intruder breaking into the home includes the following steps:

1. Opening a closed window and climbing through it.

2. Stepping on a weight sensor that is placed on the floor.

3. Leaving the room through a door within 20 seconds after entering the room.

This exact scenario is illustrated in Figure 4.1, in which both the raw input data and the described pattern are presented.

## 4.1   Notations

**Initialization Notations.**   The initialization input for our problem can be described as follows: $\mathcal{M} = INIT(E, A, O, \mathcal{D}_0, L, \textit{max-conds})$, where the events $E$, attributes $A$, and operators $O$ of interest (e.g., $=, \neq, >, <$, or any other Boolean comparison function) are provided by the domain expert and should be significant for describing critical phenomena in the data. The expert is also expected to provide the maximal length of patterns $L$, and the maximum number of conditions on a single event $\textit{max-conds}$, since they are domain-specific. The last initialization input $\mathcal{D}_0$ represents the minimal preexisting pattern knowledge (a small number of known simple patterns). Once the system is configured and provided with a minimal amount of labeled examples, we can construct patterns that aim to maximize a target reward $f$ (described in Section 5.2.2).

**Labeled Data.**   We use $\mathcal{D}_0$ to describe the group of labeled patterns given by an expert when the system is being initialized. Labels are given in the form of numerical ratings that signify how interesting the patterns are for the domain expert. During training, at predetermined points, the model will query an expert about a group of new patterns and update $\mathcal{D}_{i+1} = \mathcal{D}_i \cup new\_group$. We also annotate a group of already known labeled patterns that are not seen by the model as $\mathcal{D}'$. In our work, we use the patterns of $\mathcal{D}'$ to evaluate the generalization of solutions; meaning how well they perform on *out-of-sample* (unseen) data. In real use cases, if data is scarce or unattainable this group can be omitted without harming the learning process.

## 4.2 Metrics

We now formalize the definitions of the target metrics used throughout this work.

**Number of Queries.** One of the cornerstones of this work is the lack of experts in many applications and the high cost of expert hours. Thus, a key metric that we track for all experiments is the amount of queries sent to an expert during training. We assume all queries are equivalent in terms of difficulty, so the number of queries represents the time the system needs from the expert.

**Accuracy.** We task our model with replicating the work of a highly-skilled domain expert. To compute how well a model replicates an expert's decisions, we use the weighted accuracy metric, often referred to as the balanced accuracy [BOSB10], over both the training data $\mathcal{D}_i$ and the out-of-sample data $\mathcal{D}'$ along various points in the training process. The balanced accuracy for pattern's rank prediction is defined as follows:

$$balanced - accuracy = \frac{1}{2} \cdot \left( \frac{TP}{TP + FN} + \frac{TN}{TN + FP} \right) \qquad (4.1)$$

Where $TP$ is true positive, $TN$ is true negative, $FN$ is false negative, and $FP$ is false positive.

**Frequency.** It is not enough to mine complex and meaningful patterns. In every domain, there are infinite patterns that are theoretically interesting, but never appear in the data. Thus, in order to find patterns that are both interesting and relevant (have enough instances in the data-stream), we decided that the frequency of appearance of mined patterns is another important goal we aim to maximize.

Based upon these metrics and the notations provided in Section 4.1, **we can describe the problem of our work** as follows: how to mine frequent, meaningful patterns, when also maximize the accuracy of pattern ranking prediction over $\mathcal{D}_i$ and $\mathcal{D}'$, while minimizing the number of queries that were sent to an expert.

## 4.3 Important Considerations

We considered two major constraints to be part of the problem and essential for any solution: pattern language and action space size.

**Pattern Language.** We emphasized the importance of an easily customized pattern language, where it would be possible for an expert to define $E$, $A$ and $O$ based on the domain and to be re-defined in case changes occur in the data-stream. Hence, we aimed for a language structure that could be adapted to new domains with minimal effort while still being expressive enough to describe patterns with complex relations. The emphasis on the trade-off between customizability and expressiveness had a major impact on our approach to this problem.

While we wanted to support a generic pattern language, we had to set a limitation on conditions between two events and limit them to compare values of the same attribute (e.g., can only compare the speed attribute of event a and event b and not the speed attribute of event a and the height attribute of event b). This was done since comparing two distinct attributes is harder for experts and cannot be done trivially in most cases. For example, if an expert is given a pattern that requires the x axis speed of event a to be larger than the temperature of event b, it would take him much longer to understand than a pattern comparing the velocity of the two events.

**Action Space.** We aimed for a user-specified, rich, and expressive pattern language. However, that demand caused an increase in pattern space size and thus an increase in the size of the action space needed for pattern representation. This increase created challenges for our deep learning approach.

More formally, we use the notations defined in Section 3.2 to describe a scenario where one of $|E|$ events can be selected, and up to *max-conds* conditions are allowed from a group of $|C| = |A| \cdot |O| \cdot L$ possible conditions (since every condition has a single attribute, a single binary operation, and a single target in the comparison (which can be one of the $L - 1$ events in the pattern or a constant value), the action space would be of size:

$$|Action - Space| = |E| \cdot \sum_{i=0}^{max-conds} \binom{|C|}{i}$$

Which means that the $|Action - Space|$ is exponentially larger than $|C|$. In order to present the magnitude of this value for current day data-sets, we present the action-space size calculation for all data-sets used in our work in Table 6.1.

The authors of [TPK19, LHP$^+$19] found the problem of training discrete-action RL models over large, multidimensional actions spaces likely to be intractable. Efficient representation of this multidimensional action space is a key requirement for successful pattern mining in any user-specified CEP pattern language.

Moreover, if we look at the complete pattern space, we can see that it is exponentially larger as well. Following the notations we presented earlier in this chapter, if every pattern has between 0 and $L$ events, and every event has between 0 and $max - conds$ conditions on it, we receive the following size for the pattern space:

$$|Pattern - Space| = \sum_{i=0}^{L} \left( |E|^i \cdot \left( \sum_{j=0}^{max-conds} \binom{|C|}{j} \right)^i \right) = \sum_{i=0}^{L} |Action - Space|^i$$

Such a pattern space is huge and, as we empirically show in Section 6, attempting to search it using vanilla RL methods will produce poor results.

Tailor made solutions are needed to make it feasible for the CEP pattern generation mechanism to perform efficient exploration.

The following table contains multiple examples for action and pattern space calculations, in order to better help the reader understand their magnitude:

| $E$ | $A$ | $O$ | $L, max-conds$ | Action-Space | Pattern-Space |
|---|---|---|---|---|---|
| {A, B} | {x,y} | {<, >, =} | 5,3 | 9052 | 6.07e+19 |
| {A, B} | {x,y} | {<, >, =} | 3,5 | 25232 | 1.61e+13 |
| {A, B} | {x,y} | {<, >, = $\neq$ } | 5,3 | 21402 | 4.49e+21 |
| {A, B, C} | {x,y,z} | {<, >, =} | 5,3 | 45678 | 1.99e+23 |
| {A, B, C, D, E, F} | {x, y, z, velocity} | {<, >, =} | 5,3 | 216306 | 4.73e+26 |

It is important to note that we consider conditions that contain comparisons with a constant value as a single action (instead of one action for every constant value). Without this, both the action-space and the pattern-space are infinite. Although this seems trivial at first, solving this representation issue is a severe challenge.

# Chapter 5

# REDEEMER

## 5.1 Introduction

REDEEMER is a semi-supervised method for sequential frequent pattern mining. An overview of the system can be seen in Figure 5.1, where the major system components are presented.

REDEEMER works iteratively. At every iteration, a new window from a data-stream is inspected, embedded, and sent to the *Pattern Constructor* module, which is tasked with constructing meaningful and frequent patterns that appear in the inspected window. Later the new patterns are forwarded to the *Rank Prediction* module that evaluates how interesting patterns are to the domain expert. For this evaluation, all previous labels from prior iterations are used. If an expert is not available at that time, the best pattern according to this evaluation will be selected. If an expert is available, REDEEMER will send all the new patterns to the *Candidate Selector* module responsible for selecting what patterns to query the expert about (in this case, we update the rank prediction module according to new labels and re-evaluate our patterns). At the end of the process, the pattern constructor is updated based on its performance (rewarded for good patterns and punished for meaningless ones), and the expert receives a new mined pattern (this is further explained in Section 5.2.3).

The rest of this chapter provides an in-depth description of REDEEMER's major components, including the pattern constructor, rank predictor, and candidate selector. We also provide a thorough explanation about the overall architecture of REDEEMER.

## 5.2 Pattern Constructor

The first major component in the REDEEMER flow is the pattern constructor. The pattern constructor is responsible for parsing the input data-stream and for suggesting a pattern for the stream. In the example illustrated in Figure 4.1, the *meaningful CEP rule* was the output of the pattern constructor module.

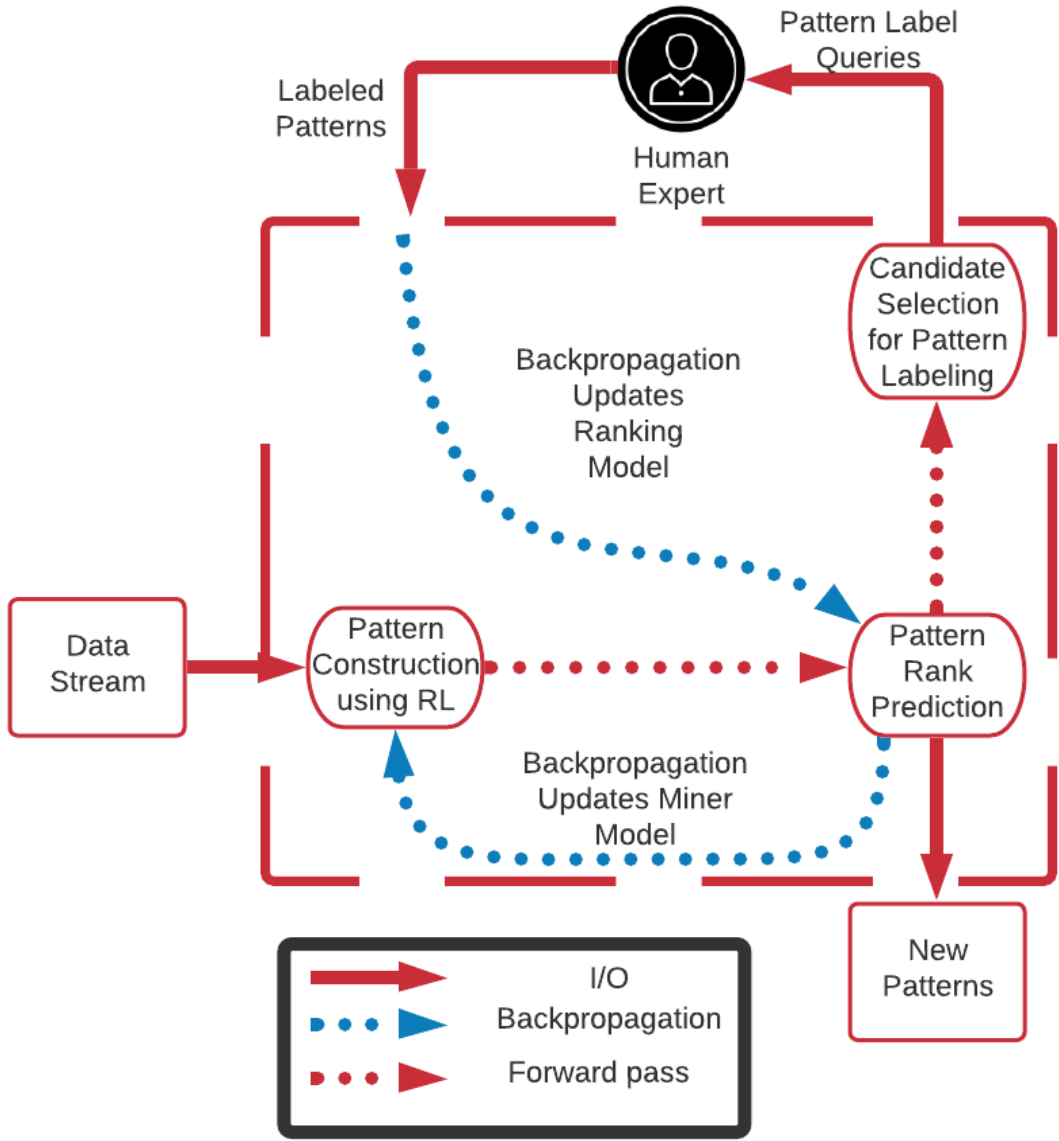

Figure 5.1: Overview of the REDEEMER training process. A data-stream is connected to the system. At every iteration, a new window for the stream is used as an input and is embedded for the reinforcement learning based pattern constructor. The pattern constructor suggests a new pattern that should be both meaningful for a domain expert and frequent in the data. The pattern rank predictor evaluates the suggested pattern and predicts its rating. If a human domain expert is present, the candidate selection module chooses patterns to query the expert about their rating. When the iteration is finished, all modules (all are neural networks) are updated, and new patterns are suggested as the system's output.

We focus on highly complex and diverse data streams, with a particular emphasis on scenarios where expert involvement is limited or nonexistent, so most supervised learning methods cannot be used, and other learning-based methods are required. We had decided on a partially supervised RL-based solution in which the agent traverses the pattern spaces, selecting one event or condition at a time.

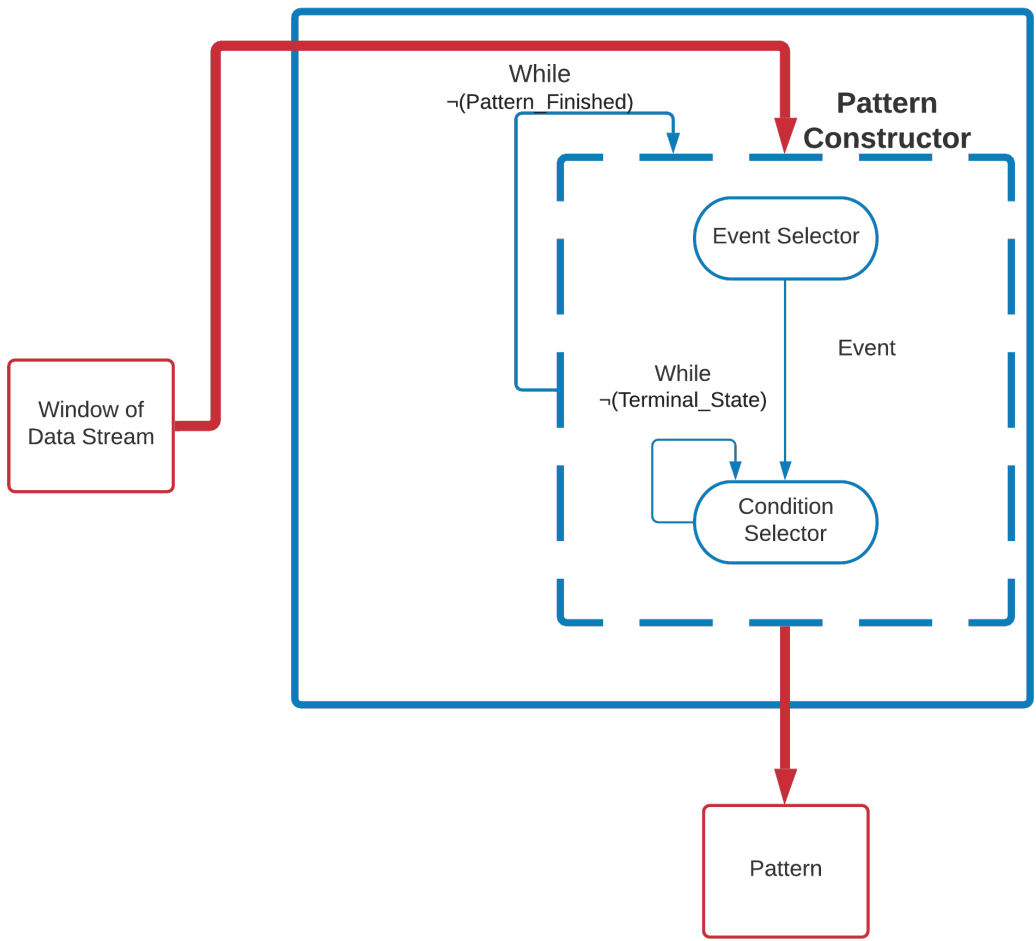

Figure 5.2: An in-depth look at the pattern constructor module. First, the window of the data-stream is embedded, and an empty pattern is created. Then, a single event is added to the pattern. Next, conditions on the new event are constructed until a terminal state is reached. Finally, the process is either repeated or finished according to the model's state.

### 5.2.1 Model Structure

We use the A2C method (described in Section 3.2) for two different tasks: event selection and condition selection. Our pattern constructor contains two networks: the *Event selector* responsible for events and the *Condition selector* responsible for selecting conditions that describe relations between previously selected events. In Figure 5.2, we illustrate the workflow between those two networks. At initialization, an empty pattern is created. Then the event network selects a new event for the pattern. Later the condition network can either add a condition for the last event or finish the last event's modification (the condition selection network reaches a terminal state). The process is repeated until the event selector reaches a terminal state.

**Terminal States**

As mentioned in Table 3.1, a *terminal-state* is a state $s \in \mathcal{S}$ from which there is nothing to do or learn from. In our work, we have multiple cases in which a terminal case can be observed.

**The event selector network** can either encounter a terminal state by selecting a *nop* (no operation) event or by receiving a zero reward for the existing pattern. The nop operation was used as part of the event selection layer (in practice, if there are $n$ events, we used a $n+1$ sized layer); this allowed the event selector to finish the pattern construction at an early stage, without completing all iterations. The zero reward terminal state means that the current constructed pattern has already no appearance in the data-stream. Every iteration of the system can be thought of as a refinement of the previous iteration. When we add more events and conditions to the existing pattern, we focus on a more specific sub-pattern of the original one. Thus, if the existing pattern is not present in the data, there is no reason to keep refining it since it would have the same reward.

**The condition selecting network** can encounter a terminal state either when it selects a *nop* condition (similar to the one described for the event selector) in all its selection layers (can be seen in blue in Figure 5.3), or if it already selected $max - conds$ conditions.

**Action Branching**

As noted in Section 4.3, we had to work with an immense multivariate action space that prior work [TPK19, LHP+19] found likely to be intractable for most RL methods. To maintain the expressiveness of our pattern language, we avoided vanilla actor-critic solutions that contain one action layer and instead chose an *action-branching* [TPK19] based solution. The leading concept in action-branching is a shared neural network module that is followed by several network branches (each branch is a different output layer). Action-branching has proved useful when facing scenarios where a combinatorial number of actions are possible, and there is a way to group actions.

To explain how action branching works, the following example is instrumental. Think of a mechanism to mimic the movement of an 8-leg octopus. If an action is chosen from the full-fledged space of potential octopus movements then the number of possibilities is very high as it consists of movement of the body plus movements of all eight legs. To reduce the potential action space, a separate leg model can be used to decide on the individual movement of each leg, and a shared, body model can be used to control all leg models and decide on the octopus body movement. Importantly, while both designs target the full space of octopus movements, the latter can use smaller models to explore smaller spaces: each leg model will explore a much smaller space of single-leg movements, and the same holds for the body model.

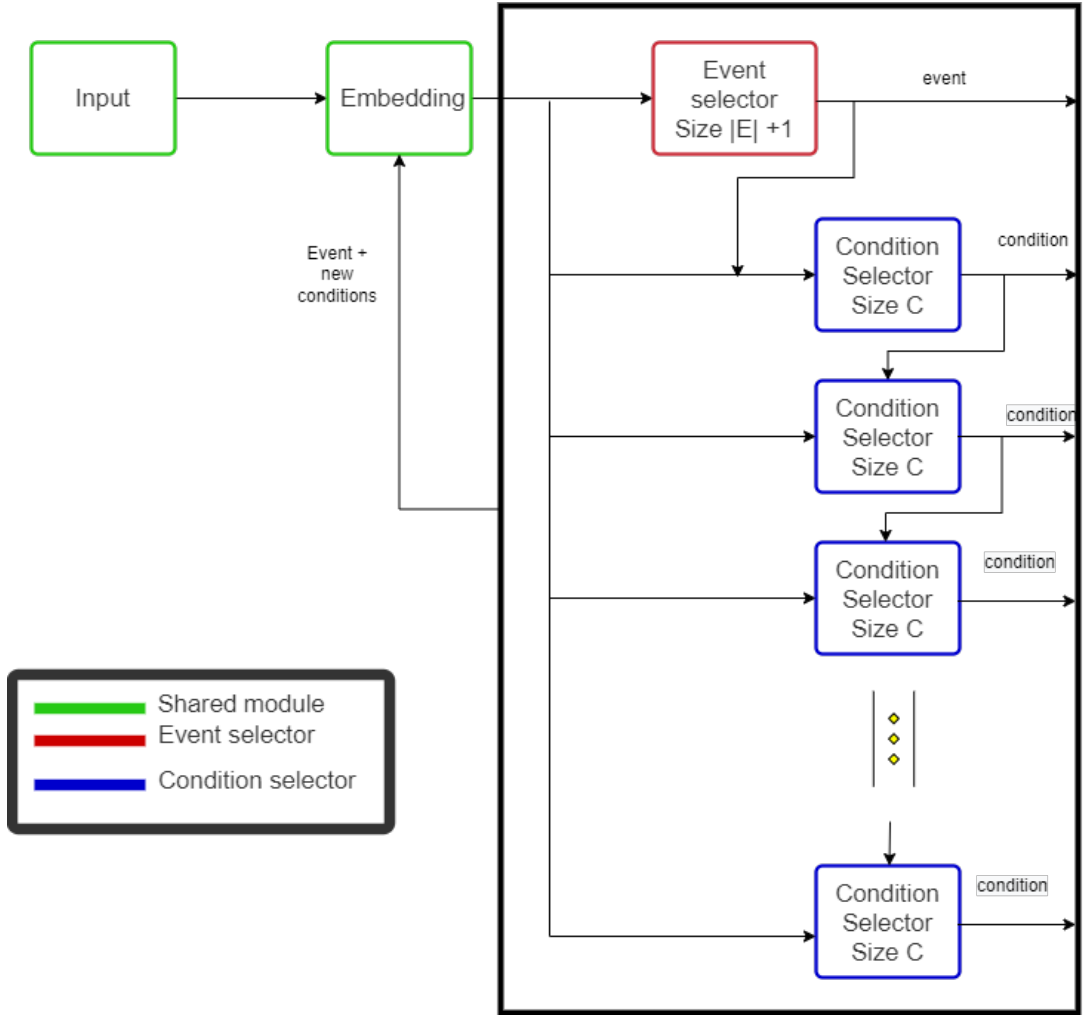

Figure 5.3: An overview of REDEEMER's pattern constructor split structure. At first, data is embedded and passes through the shared parts of the network. Later it is passed to the event selection network (which has a liner layer with an output size of |E| + 1), and its output is used to select the next event for the pattern. This decision, along with the output of the last shared layer, is used for selecting the first condition (part of the condition selection network, a linear layer with an output size of |C|). Then the second condition is selected based on the first condition and the previously mentioned information. This is repeated $max-conds$ times until all conditions are selected. Then the new event and the condition selected by both networks are used to select the next event (and the process continues).

In our case, we use a slightly different architecture: first, we have the shared module that observes the existing state (this module can be seen colored in green in Figure 5.3). Then we use the output of the shared module (embedded observed state) as the input for the event selector network, and choose a new event for the pattern. Next, we combine this selected event with the information of the shared module and connect it to our *branches* which are the condition selection layers (who are all part of the condition selecting network). Since it is possible to select up to *max-conds* conditions on every event, we have max-conds branches in the condition selecting network, each responsible for the $i^{th}$ condition.

In this way, we achieved better representation for the action space and reduced the number of actions to select from at each time. This allowed the RL agent to traverse similar trajectories easily. Moreover, our action-branching approach had significantly scaled down the size of every layer, allowing us to achieve better GPU memory usage, as we were able to load only the relevant layers every time while still representing the same extensive pattern space. An in-depth look that presents every event and condition selected in the process and illustrates the complete architecture of the pattern constructor, can be seen in Figure 5.3.

When comparing our solution to vanilla alternatives we consider, we can see that our largest layer is of size $\max(|E| + 1, |C|)$ where if we used a more naive solution in which the whole pattern is constructed at once, the largest layer would be of size $|Pattern - Size|$ (defined in Section 4.3), if we used a solution that has one layer for events and one layer for selecting conditions (selecting all conditions in a single branch) the largest layer would be of size $\max(|E| + 1, \frac{|Action - Size|}{|E| + 1})$.

The difference in size between layers sizes in the two presented pattern construction options is significant and can even be by a few orders of magnitude when working with current day data-streams.

**Policy Gradient**

Next we define the following:

- $\theta_1$: The event selecting network parameters.

- $\theta_2$: The condition selecting network parameters.

- *max-conds*: The maximum number of conditions allowed to be selected on a single event.

- $k_t$: For a given step $t$, we denote $k_t$ as the number of conditions selected on the $t^{th}$ event. By definition $k_t \in [0, max\text{-}conds]$.

- $s_t$: Represents the state before selection of the $t^{th}$ event.

- $s_{t,j}$: Represents the state after the $t^{th}$ event was selected, and $j^{th}$ conditions were selected on the $t^{th}$ event ($j \in [0, max\text{-}conds]$).

- $a_t$ and $a_{t,j}$ are defined in a similar manner for the action group. Every action performed affects the next received state.

Using those notations, we define the *combined log value function (CL)*, which is calculated as follows when selecting the $t^{th}$ event and its conditions:

$$CL(t) = \log(\pi_{\theta_1}(a_t|s_t)) + \frac{\sum_{j=0}^{k_t}\log(\pi_{\theta_2}(a_{t,j}|s_{t,j}; a_t))}{k_t} \qquad (5.1)$$

Using Formulas 3.1 and 5.1, we define a new *policy gradient*:

$$\nabla_\theta \mathcal{J}(\theta_{1,2}) = \mathbb{E}_\pi[\Sigma_{t=0}^{T-1}\nabla_{\theta_{1,2}}CL(t)A(s_t, a_t)] \qquad (5.2)$$

Our combined log value function replaces the *expected future reward* ($G_t$) that is used in vanilla A2C (Equation (3.1)). The combined log value is different from $G_t$ since it considers both the event and the condition networks and combines their impact on the reward received for a new mined pattern with the same magnitude (both have the equal effect on the result). This directs the agent to always improve on both fronts: find better events and select better conditions for those events to generate more complex, interesting, and frequent patterns.

Our policy gradient unifies the two distinct networks towards the same goal. A pattern constructed can only be meaningful if both its events and conditions together describe an interesting situation. Therefore, it is crucial that both the event and condition networks are trained simultaneously (similar to the method presented in [GPAM+14] that is used to train generative adversarial networks (GANs)). In this way, both networks improve gradually while always performing at almost the same level.

Both $\pi_{\theta_1}$ and $\pi_{\theta_2}$ are computed using UCB1, meaning that actions are selected according to Equation (3.2). This was important for improving the exploration of the pattern space and will be further explained in Section 5.6.

A detailed representation of our A2C variation on a single learning episode [1] can be seen in Algorithm 5.1.

### 5.2.2   Reward Function

It was important for us to find a balance between the frequency and relevance of patterns. In the smart home scenario described in Figure 4.1, we note the pattern of a home invader as both frequent (enough to be noticed in the context of a large home security company) and meaningful. Other, less valuable patterns may also be presented. For example, a pattern of a door opening would be considered frequent but not meaningful. And a pattern describing a mythical being climbing down the chimney might be extremely interesting but not frequent enough to be understood and exploited by an expert. To mine patterns that fit this criterion, we constructed a reward function based on both frequency of appearance and the usefulness of the pattern to a domain

---

[1] sequence of actions and states from the starting state until the terminal state.

**Algorithm 5.1** REDEEMER A2C single episode algorithm.

1: Create an empty pattern P
2: **for** $t = 1...$T and $\neg($ terminal $s_t)$ **do**
3:     Select $a_t$ according to policy $\pi_{\theta_1}(a_t|s_t;\theta_1)$ and add corresponding event to P
4:
5:     **for** $j = 1...k_t$ and $\neg$ (terminal $s_{t,j}$) **do**
6:         Select $a_{t,j}$ according to policy $\pi_{\theta_2}(a_{t,j}|s_{t,j};a_t;\theta_2)$ and add corresponding condition to last event in P
7:         Receive new state $s_{t,j+1}$
8:     **end for**
9:     Receive new state $s_{t+1}$ and reward $r_t$
10:     Store value function in $V_t$
11:     Calculate $CL_t$ according to Equation (5.1)
12:     $lastT \leftarrow t$
13: **end for**
14: R $\leftarrow \begin{cases} 0 & \text{for terminal } s_{lastT} \\ V_{s_{lastT+1}}(\theta_{1,2}) & \text{for non-terminal } s_{lastT} \end{cases}$
15:
16: **for** $t = lastT...1$ **do**
17:     $R \leftarrow r_t + \gamma * R$
18:     $R_t \leftarrow R$
19:     $A_t \leftarrow R_t - V_t$
20: **end for**
21: Update $\theta_1, \theta_2$ according to Equation (5.2)-actor and MSE-critic.

expert.

All the used notations and symbols for pattern reward are summarized in Table 5.1.

| symbol | definition |
| --- | --- |
| *Rew* | The reward function of a pattern. |
| *freq* | Frequency based function, which approximated the appearance frequency of a pattern over the data-stream. |
| *rating* | The rating of a pattern, represent the importance of the pattern to the domain expert, can be either given by the expert or estimated by REDEEMR. Further explained in Section 5.3. |
| $appearances(pattern, window)$ | The number of appearances of the given pattern in the given time-window (number of pattern matches). |
| $window_i$ | The $i^{th}$ received time-window in the data-stream. |
| $jump - interval$ | Value unique for each data-set. The jump interval is an integer the determines what previous time-windows are relevant for approximation pattern frequency in the past. Determines what is the offset between time-window indexes |

Table 5.1: Symbols for pattern's reward calculation.

Using these notations, we can define the pattern reward function *Rew* to be calculated as follows:

$$Rew(pattern, time-window) = freq(pattern, time-window) \cdot rating(pattern) \quad (5.3)$$

When working with high-speed online data-streams, calculating the appearance frequency is either impossible because the whole data-stream is never stored at once as it is often done in online setups, or impractical because it is too time consuming. Our method approximates the frequency of appearance by averaging results on near windows in the data-stream.

When given a time-window of data-stream with index $i$, the calculation is as follows:

$$freq(pattern, window_i) = appearances(pattern, window_i) \cdot 0.5+$$
$$appearances(pattern, window_{(i-jump\_interval)}) \cdot 0.25+$$
$$appearances(pattern, window_{(i-2 \cdot jump\_interval)}) \cdot 0.25$$

This adjacent window-based method allows our model to adapt to drastic changes in the data stream, this can be seen in following example. Following the example seen in Figure 4.1, if the Door1 sensor malfunctions and does not transmit information after a certain point, the reward for patterns containing this sensor will slowly fade, and the model would not suggest those patterns after enough time passes. On the other hand, if the same sensor malfunctions momentarily and returns to work shortly after, the rewards that are based on previous time-windows would prevent the reward from zeroing.

The pattern's usefulness (rating in Equation (5.3)) is either given by the domain expert or approximated by the rank predictor (described in Section 5.3) using previously supplied labels.

### 5.2.3   Model's Output

Although we use the combination of frequency and ranking (usefulness) as components in reward calculation for the RL process, we maintain both parts separately for every pattern. When the training is finished, we use the *Pareto efficiency* to select the best patterns we mine during the training process and submit them to the system's user.

The Pareto efficiency describes a situation where there are multiple objectives (criteriums, goals), and every solution can only improve some of those while harming others (a solution that maximizes all objectives can not be found). We used the pattern's frequency of appearances and rating as the two distinct objectives for the Pareto efficiency selection, and chose results according to their value w.r.t. this evaluation.

## 5.3 Rank Predictor and Candidate Selector

Although they are presented as two distinct components, the rank predictor and candidate selector work together throughout the learning process. Both of these modules aim to replicate the expert's ability to distinguish between meaningful and meaningless patterns. Their ranking is also used as part of the reward of the pattern constructor and thus guides its exploration of the pattern spaces.

To achieve this and enable REDEEMER to utilize the limited expert availability to its fullest capacity, REDEEMER uses an active learning model for expert interaction. Predetermined interaction points are designated, and when reached, the candidate selector queries the expert about new mined patterns. At any interaction point, the model is given a range for the number of queries allowed and is encouraged to minimize the number of queries and achieve high accuracy in rank prediction. The decision about the number of queries and the specific candidates selected considers past success trends and the uniqueness of patterns it observed since the last expert interaction. A simplified overview of this process can be seen in Algorithm 5.2.

### 5.3.1 Rank Predictor

After a pattern has been constructed and its frequency was estimated, it is sent to the rank predictor, who evaluates the relevance and importance of the pattern for the application domain. The rank predictor utilizes the most updated $\mathcal{D}_i$ to determine a pattern's rank.

---

**Algorithm 5.2** REDEEMER interaction with an expert.

**Require:** $\mathcal{N}$ new patterns mined, $[x_1, x_2]$ given range and $\mathcal{M}$ a REDEEMER instance with existing labeled data $\mathcal{D}_i$.

1: **if** Interaction point is reached **then**
2:     Select $n \in [x_1, x_2]$ for the number of queries to the expert.
3:     Select n patterns from the $\mathcal{N}$ given patterns, using the uncertainty based strategy mentioned in Section 5.3.2.
4:     Query the expert on the n patterns, receive labels, and mark them as *new_group*.
5:     Compute $\mathcal{D}_{(i+1)} = \mathcal{D}_i \cup new\_group$.
6: **end if**
7: Return $\mathcal{M}$

---

**Challenges in rank prediction** Our rank predictor faced a difficult classification problem because its initial training set does not cover the entire pattern space. Moreover, since the pattern constructor is encouraged to further explore the pattern space (described in Section 5.6), the predictor is tasked with predicting ranking for patterns that are very dissimilar to those it has seen before; it needs to continuously adapt to changes in the incoming examples and to the ever-changing labeled set. To solve those problems, we employed an active learning based approach (incremental training over the changing data-set) that was controlled by the candidate selector.

### 5.3.2 Candidate Selector

Since the size of our pattern space is vast and under the assumption of minimal knowledge available before training, the need for more labels is almost inevitable. The limited availability of a domain expert led us to search for methods that better utilize the expert's time. For candidate selection, we used a batched confidence-based strategy, also referred to as an uncertainty-based strategy [RXC+21]. In this strategy, when the candidate selector was given the option to send queries to the expert, all candidate rankings were re-evaluated. Those to whom the selector gave the lowest confidence in their value prediction were sent for labeling by the expert.

After an interaction point with the expert has been reached and response for queries is received, the candidate selector sends the expert's response to the rank predictor to update $\mathcal{D}_i$ with the received labels. Later the rank predictor's weights will be updated according to its mistakes in prediction (w.r.t. the received ground truth labels). This is also important for the next interaction point, to re-evaluate the areas in the pattern spaces that still have high uncertainty in their rank prediction. A simplified overview of this process can be seen in Algorithm 5.3.

---

**Algorithm 5.3** Candidate Selector

---

**Require:** $\mathcal{N}$ new patterns mined, max_size ms (maximum number of patterns from same rank allowed) and $\mathcal{M}$ a REDEEMER instance.

1: Run $\mathcal{M's}$ rank predictor on all new mined patterns and obtain their rank and certainty.
2: Sort all patterns by certainty.
3: set count[i] = 0 $\forall i \in$ ranks.
4: **for** pattern $p$ in all new Patterns **do**
5:     get $r$ the predicted rank of $p$.
6:     **if** count[r] > max_size **then**
7:         set $p$ in the end of the pattern list.
8:     **end if**
9: **end for**
10: Return list of pattern ordered by their uncertainty.

---

## 5.4 Pattern Language

We followed the pattern construction method suggested by the authors of [BDO19] that splits patterns into Conditions Tree and Events Tree. We limited the pattern languages to only include sequential patterns occurring in a user-specified time interval (since the pattern space is already vast in size, we chose to focus on mining sequential patterns). In our condition tree, we allowed the selection of multi-ple conditions on a single event; these conditions define requirements between attributes of different events or between an attribute and a numerical value. For example, when experimenting with the Football dataset [MZJ13], every condition contained one of the following operators $\{>, <, =, \not>, \not<, \neq\}$ between one event's attribute $\{x\_loc, y\_loc, z\_loc, x\_v, y\_v, z\_v, x\_acc, y\_acc, z\_acc\}$ and either a numerical value or another event's attribute. Conditions were separated by $\vee$ or $\wedge$. Examples of patterns generated in our experiments over the Football dataset can be seen in Table 5.2.

| Events & Time | Conditions | Explanation |
|---|---|---|
| {Player 53 left leg as a, Ball 4 as b, Ball 4 as c} where{SEQ(a,b,c)} within 0.5s | $(a.x\_loc = b.x\_loc)$ $\wedge(a.y\_loc = b.y\_loc)$ $\wedge(b.x\_loc < c.x\_loc)$ | A player touches the ball with its left leg. The ball rolls away along the x axis of the field. |
| {Ball 4 as a, Player 47 left leg as b Player 88 right leg as c, ball 4 as d}, where{SEQ(a,b,c,d)} within 2s | $(a.x\_loc <$ $33762 \wedge a.x\_v > b.x\_v)$ $\wedge(b.x\_loc > 34000)$ $\wedge(c.x\_loc > 34015)$ $\wedge(d.x\_loc >$ $34800 \vee d.x\_v = 0)$ | The ball is on the left side of the field (middle x_loc is 33960-33965) while being faster than player 47.<br><br>Players 47 and 88 are beyond the middle of the field. The ball either crosses the middle or stops. |

Table 5.2: Patterns generated from the Football dataset.

| Events | Pattern Formula | Complete Pattern |
|---|---|---|
| {B as a, C as b} where {SEQ(a,b)} within 10s | $b.value = x_1$ | $b.value = 2048$ |
| {A as a, A as b, B as c, C as d} where {SEQ(a,b,c,d)} within 10s | $a.value < b.value$ $\wedge b.value \leq x_1$ $\wedge \neg(c.value = x_2)$ | $a.value < b.value$ $\wedge b.value \leq 1912$ $\wedge \neg(c.value = 4137)$ |

Table 5.3: Patterns and formulas. Examples of patterns with variables missing and their completed counterparts. Each $x_i$ represents the $i^{th}$ missing value in the pattern formula. Variables are completed (replaced by constant values) using Bayesian search.

## 5.5   Bayesian Search

While most of the pattern structure is constructed by the REDEEMER pattern constructor module, some values in the pattern are completed at a later stage. The RL module output can be viewed as a *pattern formula* with variables not yet found; this can be seen in Table 5.3. We used Bayesian search to complete the formula and generate a complete pattern. Value selection was done by guiding the search to maximize the frequency of the pattern's appearance in the data.

**Bayesian search** is a method for finding optimal values or locating objects in a vast search area. The Bayesian search is built upon Bayesian inference and Gaussian process, which attempts to find the maximum value of an unknown function. The search works by constructing a posterior distribution of functions that describe the function of interest (in our case, the number of appearances of the pattern in the datastream). As we compute this function for more examples, our posterior distribution improves. The algorithm becomes more certain which regions in parameter space are worth exploring and which are not.

In order to find values for our variables, we run the Bayesian search for multiple iterations. In every iteration we sample a few possible values from the current distribution estimation of the parameter space and calculate their reward by Equation (5.3), then we update the parameter space estimation accordingly. For optimal results (best possible suggested values) we can run this search until no more improvement can be found, and then select the values that resulted with highest reward.

## 5.6   Exploration-Exploitation

To efficiently explore our vast action space, we applied multiple exploration vs. exploitation techniques.

We used UCB1 to encourage the REDEEMER pattern construction model to select rarely exploited actions and further explore the available pattern space. To achieve this, we applied the UCB1 formula Equation (3.2) to re-weight every possible action (both for events and for conditions). Our process was as follows:

1. We received action scores as part of the *actor's* forward pass.

2. We calculated every action UCB score.

3. We normalized UCB scores to be between 0 and 1 (since action scores are probabilities and are also between 0 and 1) then divided the score by 2.

4. We summed the original score and the UCB value

5. We normalized scores again to all be probabilities between 0 and 1.

This caused a large increase in the score of rarely exploited actions and increased their likelihood of being selected (since lesser explored events and conditions probability of selection rose higher over time).

Our ability to adjust to changes in the data-stream and reduce the effects of inaccurate rewards obtained at the start of the learning process were both enhanced by UCB. As the data stream is expected to change over time, events and conditions that resulted in low rewards in the past could become relevant over time. Because of their UCB score, we must retry those actions and gain new rewards, enabling us to update our evaluation of them. Similarly, actions that were given low rewards in the early stages of the learning process due to inaccurate rank predictors will also gain more accurate rankings due to their increased UCB score.

Furthermore, we employed a dynamic learning rate that is reduced when the difference in probability of the selection of actions increases (i.e., when some actions are almost always selected).

Combining these methods resulted in a greater variety of unique patterns and better patterns mining according to the Pareto efficiency.

We considered the pattern's frequency of appearances and rating as two distinct objectives. We applied Pareto efficiency to find patterns that were optimal w.r.t. both objectives. In practice, all mined patterns are compared by both metrics and are ranked based on their difference from the maximum possible value for each metric.

# Chapter 6

# Experimental Evaluation

Our experimental study focused on the following four questions:

1. Quality of generated patterns: How does REDEEMER compare with other pattern mining solutions in terms of expected reward (pattern constructor module) and accuracy in rank prediction?

2. Availability problem: How well does REDEEMER perform when used with different types of experts who differ in their availability?

3. Reliability problem: How well does REDEEMER perform when used with experts who have different levels of expertise? Can REDEEMER achieve sufficient results when labels are noisy and inaccurate?

4. Granularity problem: How does REDEEMER perform when the granularity of labels from the experts is changed to more aptly describe the correctness of the pattern?

Question 1 is the main research question of this study and aims to evaluate how well our novel approach generates new patterns on several datasets. Questions 2, 3, and 4 address the problem of how to correctly utilize an expert's knowledge to achieve the best possible results and extend pattern knowledge when real-life limitations affect the training process of REDEEMER.

## 6.1 Experimental Setup

### 6.1.1 Experimental environment

We recorded all experiments using the Titan Xp GPU and 12 CPUs (Intel(R) Xeon(R) CPU E5-2680 v3 @ 2.50GHz). For every run in every experiment, we ran the model on three fixed seeds ($[0, 10, 50]$) and either presented the average of those results or presented the mean and standard deviation computed over the runs.

In all experiments we ran epochs [1] that contained 500 episodes each. In every example the input for REDEEMER was a single time-window of the data-stream

## 6.1.2   Implementation Details

### Rank Predictor

To face the challenges described in Section 5.3.1, we used another neural network based architecture for rank prediction. Unlike in our reinforcement learning module, for the rank predictor, a simple multilayered classifier proved sufficient. We used 4 linear layers with ReLU, or Leaky ReLU applied between them while also using a dropout layer after the first linear one. Finally, we used a softmax over the result of the final linear layer to predict the pattern rating. This model (the number of layers, the specific activation layers between them, and the usage of dropout) was empirically selected after performing extensive experiments over a large number of different architectures and different values for each hyperparameter of the model.[2]

### Candidate Selector

We used the output of the last layer of the rank predictor for certainty evaluation. After we applied SoftMax on this output and selected the pattern's rank, we checked and stored the value given to the selected rank. We used the SoftMax function in order to convert the linear layer values into probability for selecting every rank and stored the probability of the chosen rank. This value was used for certainty evaluation (where a value close to 1 meant high certainty).

### Bayesian Search Hyperparameters

We used CEP tools to find matches of patterns in the data-stream and to evaluate their frequency of appearance. Both Bayesian search implementation [Nog20] and CEP [Kol21] tools are publicly available packages.

Our Bayesian search process is as follows:

1. An iteration is finished, and a pattern formula is suggested by the pattern constructor. The pattern formula has some missing variables. For example, in the first row of Table 5.3, the comparison value of the $b.x$ event is missing.

2. We initialize a Bayesian search that receives a list of variables and a range of values to select from.

---

[1] An epoch is an iteration over the complete training set, the number of epochs is the number of iterations.

[2] These experiments were conducted using W&B([Bie20]) "sweeping" tool, where we tried to optimized the average results over all data-sets and across multiple seeds.

(a) The Bayesian search suggests 3 sets of values for the variables (this is done by sampling values out of the current evaluated distribution of the parameters space).

(b) The now completed value is sent to the CEP tool, which searches for matches of the pattern in the data-stream. A score for frequency is obtained and sent to the Bayesian search to improve the next iteration.

(c) The process is repeated for a predetermined number of iterations or until no improvement is found for a large number of consecutive iterations.

3. The values that resulted in the best frequency are used for the complete pattern.

We ran the aforementioned schema for 10 iterations and stopped after 5 consecutive iterations without improvement. Those values were chosen under constraints on the run time of the Bayesian search, as we did not want the value selection to be the bottleneck of the mining process. Values were empirically selected selection under the mentioned constraint and can be changed when used in other setups.

### 6.1.3 Considered Methods

We compared REDEEMER with two existing reinforcement learning methods:

- **VAC**, which is a vanilla actor-critic implementation that is widely used. VAC also had access to a standard active learning model for equal comparison.

- **DDPG**, the acclaimed Deep Deterministic Policy Gradients presented in [LHP⁺19], is an actor-critic, model-free method. We used the publicly available implementation of [HRE⁺18], which is optimized for working on large discrete action spaces. This implementation is based both on the original model presented in [LHP⁺19] and the improvements suggested in [DAEvH⁺16] (mainly the *Wolpertinger* training method, in which while training exploration noise is added to find valid action candidates). DDPG also had access to a standard active learning model for equal comparison.

We also compare our suggested REDEEMER model with two variations of it that are operated with experts who differ in their availability:

- **no_knowledge** is a variant of REDEEMER in which $\mathcal{D}_0 = \phi$, meaning the expert is not able to supply examples during initialization but can rank new patterns in a limited capacity.

- **always_available** is a baseline method in which an expert is always available and is able to rank every pattern generated. This method can be seen as an upper bound for REDEEMER since its rank prediction module is replaced with an oracle.

### 6.1.4 Datasets

We evaluated our methods on the three datasets described below [3].

| Data-set | #events | #attributes | Action-Space size |
|---|---|---|---|
| Football [MZJ13] | 41 | 9 | 4407341945 |
| StarPilot [CHHS20] | 24 | 6 | 1493006424 |
| GPU Cluster | 15 | 10 | 4032014014 |

Table 6.1: Characteristics of datasets and action space sizes (calculation based on Section 4.3).

**StarPilot.** StarPilot is one of 16 unique procgen benchmarks released by *OpenAI* and published in [CHHS20]. We adapted the publicly available source code of the *StarPilot* game to generate data-streams from game instances. To achieve this goal, we modified all game objects so they would track information about themselves (e.g., location, speed, etc.) and send it to a log file every 5 game frames.

We later trained an agent to play the game using the *procgen-train* library (also publicly available), with longevity being the sole goal of the agent. This resulted in longer games and provided us with massive data-streams.

In our modified game, different objects track different attributes. For example, the player's spaceship tracks information about its speed, location, and health points, while the finish only tracks information about its location. This resulted in a highly complex, multivariate, and large data-stream.

**Football dataset.** The Football dataset was taken from the DEBS 2013 Grand Challenge [MZJ13]. The data was collected from sensors embedded in the shoes and gloves of football players, referees, goalkeepers, and the ball itself. The data spans the entire game's duration, where each row in the dataset contains a unique identifier that represents one sensor. Each row also contains a timestamp, the location coordinates, the velocity, and the acceleration of the sensor in all three dimensions.

In the original dataset, the ball's sensor produces data at a frequency of 2000Hz, while every other sensor produces data at a frequency of 200Hz. To create patterns that are meaningful to a human expert, modifications to the original dataset were made [4]. Since we aim for a simpler and more understandable data-set, we chose to downsample the original data. Because we are working with experts that have limited availability and also cannot answer queries about patterns that occur in a time frame humans cannot comprehend, we had to downsample our data, so that information tracked by the same sensor would be closer to one that is observable by an expert watching the game.

---

[3] All data-sets are available at https://github.com/Guy-Shapira/REDEEMER.

[4] Our adapted version of the dataset is available in: https://github.com/Guy-Shapira/REDEEMER

**GPU Cluster.** The GPU Cluster dataset is a dataset that tracks information about servers that are used as part of a GPU computational cluster. Data was collected by monitoring all GPUs in a research server cluster. For over a month, the following attributes were tracked: temperature, power consumption, and memory usage in a periodic manner. The cluster contains 5 servers with identical hardware: 8 GeForce RTX 2080 Ti GPUs, and 80 Intel(R) Xeon(R) Gold 6230 CPU @ 2.10GHz CPUs.

Each entry contains an identifier of a single server and information about its 8 GPUs. To enrich our events group, we used different event types when the server was used with different load levels. For example, for server number 2, if only one GPU were used, the event identifier was called *server_2-low*, and if all GPUs were used, the event identifier was called *server_2-high*.

We used only part of the attributes tracked during the month-long gathering of data. Even with our action branching method, using all the information was impossible due to memory and run-time limitations.

For all data-sets, we gathered a few groups of patterns relevant to the application domain and could have been easily found by an expert. For example, in the StarPilot dataset, one of the patterns describes the player crossing the finish line.

### 6.1.5  Initialization Patterns

Next, we describe patterns we used as part of $\mathcal{D}_0$ for every data-set. We consider those patterns as simple patterns that could have easily been observed by a domain expert and are not hard to discover. In all cases, we also used randomly sampled patterns (sample using the uniform distribution for every event and condition) to obtain rankings for sub-par patterns (if we only provided the model with good patterns in initialization, it would have overestimated every pattern since its training set is biased). We define the following notations for simplicity and better readability of the presented patterns:

| Notation | Meaning |
|---|---|
| EqualF(a,b) (Equal location for the Football data-set) | $(a.x\_loc = b.x\_loc \wedge a.y\_loc = b.y\_loc \wedge a.z\_loc = b.z\_loc)$ |
| EqualS(a,b) (Equal location for the StarPilot data-set) | $(a.x\_loc = b.x\_loc \wedge a.y\_loc = b.y\_loc)$ |
| GPU.MU (for the GPU cluster data-set) | Memory Usage of GPU |

We present all the patterns used for the Football data-set in Table 6.2, the StarPilot data-set in Table 6.3, and for the GPU cluster data-set in Table 6.4.

| Pattern group | Events | Conditions | Explanation |
|---|---|---|---|
| Ball touch | {Player 13 left leg as a, Ball 4 as b, Ball 4 as c} where{SEQ(a,b,c)} within 0.5s | $EqualF(a,b)$ $\wedge(b.y\_loc = c.y\_loc)$ $\wedge(a.x\_loc \neq c.x\_loc)$ | Player with sensor 13 on its left leg touches the ball (same x,y and z coordinates), later the ball is in a different location. |
|  | {Ball 8 as a, Player 97 left arm as b, Ball 8 as c } where{SEQ(a,b,c)} within 0.5s | $(a.z\_loc > b.z\_loc)$ $\wedge EqualF(b,c)$ | The ball moves toward the goalkeeper's arm, and then hits it (alternatively the keeper moves towards the ball and hit it). |
| Dribble | {Player 88 right leg as a, Ball 4 as b, Player 88 right leg as c, Ball 4 as d} where{SEQ(a,b,c,d)} within 0.5s | $EqualF(a,b) \wedge$ $(a.x\_loc < c.x\_loc) \wedge$ $EqualF(c,d)$ | The player with sensor 88 on its right leg start with the ball at the same location, later he advances up the pitch, with the ball still touching his leg. |

Table 6.2: Initialization patterns ($\mathcal{D}_0$) for the Football data-set.

## 6.2 Results

### 6.2.1 Quality of Generated Patterns

Figures 6.1 and 6.2 present the comparison of baseline methods **DDPG** and **VAC** with our presented **REDEEMER** model. We ran all models over three fixed random seeds in all of the following runs in this section. In order to evaluate the performance of the rank predictor module (which is responsible for replicating the expert's ranking ability for unseen patterns), we compared the accuracy of all models on both the training data-set (i.e., the most updated $\mathcal{D}_i$) and on the out-of-sample data (i.e., $\mathcal{D}'$). The results are shown over the StarPilot dataset (Figure 6.1a), the Football dataset (Figure 6.1b), and the GPU cluster data-set (Figure 6.1c).

To evaluate our pattern constructor ability to generate meaningful patterns, we compared the weighted reward of all methods described in Section 6.1.3. We computed the weighted reward of patterns after every epoch, with the reward being calculated as mentioned in Section 5.2.2. To create a fair and meaningful comparison, we used the ground truth rating of a pattern instead of the predicted rating given by the model.

| Pattern group | Events | Conditions | Explanation |
|---|---|---|---|
| Shot a lot | {Player spaceship as a, Player bullet1 as b, Player bullet2 as c, Player bullet3 as d} where{SEQ(a,b,c,d)} within 0.5s | $EqualS(a,b) \wedge$ $EqualS(a,c) \wedge$ $EqualS(a,d)$ | The player spaceship shoots at least 3 different bullets (every bullet independently) while remaining at the same location. |
| Finish line | {Player spaceship as a, Enemy spaceship1 as b, Player spaceship as c, Finish line as c} where{SEQ(a,b,c,d)} within 1s | $(a.x\_loc <$ $c.x\_loc) \wedge$ $EqualS(c,d)$ | The player's spaceship moves towards the finish line and crosses it. At the same time, an enemy spaceship is observable on screen. |

Table 6.3: Initialization patterns ($\mathcal{D}_0$) for the StarPilot data-set.

| Pattern group | Events | Conditions | Explanation |
|---|---|---|---|
| Load increase | {Server 3 low as a, Server 3 reg as b} where{SEQ(a,b)} within 2s | $a.GPU2\_MU =$ $0 \wedge$ $b.GPU2\_MU > 0$ | Load in server 3 increases as GPU 2 in server 3 memory usage gets used. |
| Raise and Drop | {Server 4 reg as a, Server 4 high as b, Server 4 reg as c} where{SEQ(a,b,c)} within 3s | $a.GPU4\_MU \neq$ $0 \vee$ $a.GPU3\_MU \neq 0$ | Load over server 4 increase from regular load to high one, but quickly drops back to regular, while at least one of GPUs 3 or 4 are used. |

Table 6.4: Initialization patterns ($\mathcal{D}_0$) for the GPU cluster data-set.

In this way, it would not be biased in favor of non-accurate models that overestimated rankings. Our results over the StarPilot data-set can be seen in Figure 6.2a, over the Football dataset in Figure 6.2b, and over the GPU cluster dataset in Figure 6.2c.

To compare other methods with the VAC on equal footing; we had to use a less expressive pattern language. We limited comparisons of two attributes to only be between succeeding events and reduced the limit for the number of conditions on a single event (*max-conds*). These changes allowed the baseline methods to run on identical hardware. In all other experiments, we used the same, more complex pattern language[5].

The results show that both baseline methods do not converge on a sufficient solution and do not efficiently explore the vast action space while REDEEMER can overcome these problems. The baseline methods were generally inferior to REDEEMER in all datasets used in this experiment, and they did not show a clear improvement trend either. In Figure 6.1a, an unexpected anomaly occurred when DDPG's accuracy over $\mathcal{D}_1$ was the highest accuracy measured out of all models during the training process. This phenomenon can be explained by the model being quick to learn rank prediction for the small and similar initial pattern group, but as the later results show, its performance quickly dropped when it encountered new patterns.

It is important to note that some of the models achieved better out-of-sample accuracy than the in-sample accuracy. Although very unusual for most deep learning models, in our case it was almost unavoidable. The out-of-sample data $\mathcal{D}'$ is a fixed group of patterns while $\mathcal{D}_i$ is a dynamic ever-changing set of patterns that is extended based on the patterns generated by the patterns constructor. Many times during the learning process, the model was tasked with evaluating rankings for patterns that were completely new and different from those that were seen in previous iterations of REDEEMER. For those patterns the model often predicted completely incorrect values until a sufficient number of labels were given by the expert for similar patterns (this process can spread across multiple hours of run-time). This caused for drops in in-sample accuracy (e.g., in Figure 6.2b VAC's accuracy drops dramatically after 10 hours and after 20 hours of training).

From Figure 6.2, it is clear that the baseline methods do not manage to produce a significant improvement during the long training phase of 50 epochs (upward of 80 hours). This can be attributed to the challenges mentioned in Section 5.2.1.

Finally, in Table 6.5, we present an example pattern mined for each data-set. Those 3 patterns were mined by REDEEMER in 1 of the 3 seeds used and were selected as one of the top-10 patterns mined during the learning process. Between those 30 examples, we chose 3 representative examples that were both distinct from the patterns provided in $\mathcal{D}_0$ and easy to understand without vast expertise.

---

[5]This was only applied to the Football dataset, which has the largest action space.

| Data-set | Events | Conditions | Suggested Explanation |
|---|---|---|---|
| Football | {Player 88 right leg as a, Ball 4 as b, Player 58 right leg as c, Ball 4 as d, Player 88 right leg as e, Ball 4 as f} where {SEQ(a,b,c,d,e,f)} within 2.5s | $EqualF(a,b) \wedge$ $(a.x\_loc \neq c.x\_loc \vee$ $a.y\_loc \neq c.y\_loc) \wedge$ $EqualF(c,d) \wedge c.x\_v >$ $0 \wedge EqualF(e,f)$ | Player with sensor 88 on his right leg starts with the ball, he passes the ball to the player with sensor 58 on his right leg (teammate) who is at different location and is not stationary (x axis velocity is not zero), and later receives the ball back (known as double-pass). |
| StarPilot | {Player as a, Bullet1 as b, Bullet2 as c, Enemy spaceship1 as d, Bullet1 as e, Enemy spaceship2 as f, Explosion1 as g, Explosion2 as h} where {SEQ(a,b,c,d,e,f,g,h)} within 3s | $(a.health\_points =$ $3 \wedge EqualS(a,b)) \wedge$ $EqualS(b,c) \wedge$ $EqualS(e,d) \wedge$ $Equals(d,g) \wedge$ $(f.health\_points =$ $0 \wedge EqualS(f,h)$ | The player starts with 3 health points. He shoots 2 bullets without moving. The first bullet (events b and e) hits an enemy spaceship and an explosion animation appears at the same place. Later another enemy spaceship has zero health points (destroyed) and an explosion animation appears at the same place (presumably hit by the second bullet shot by the player). |
| GPU-Cluster | {Server 4 low as a, Server 4 low as b, Server 4 reg as c, server 4 reg as d} where{SEQ(a,b,c,d)} within 10s | $a.GPU4\_MU =$ $0 \wedge b.GPU4\_MU >$ $a.GPU4\_MU \wedge$ $c.GPU4\_MU >$ $b.GPU4\_MU \wedge$ $d.GPU4\_MU >$ $c.GPU4\_MU$ | Load over server 4 start as low, and GPU 4 is not utilized. Then there is an increase of the memory used by the GPU along two consecutive samples, and the server's status changes from low to regular. Finally the memory usages stabilize in event d. |

Table 6.5: Examples for patterns mined by REDEEMER during the training process of experiment 1 (Section 6.2.1).

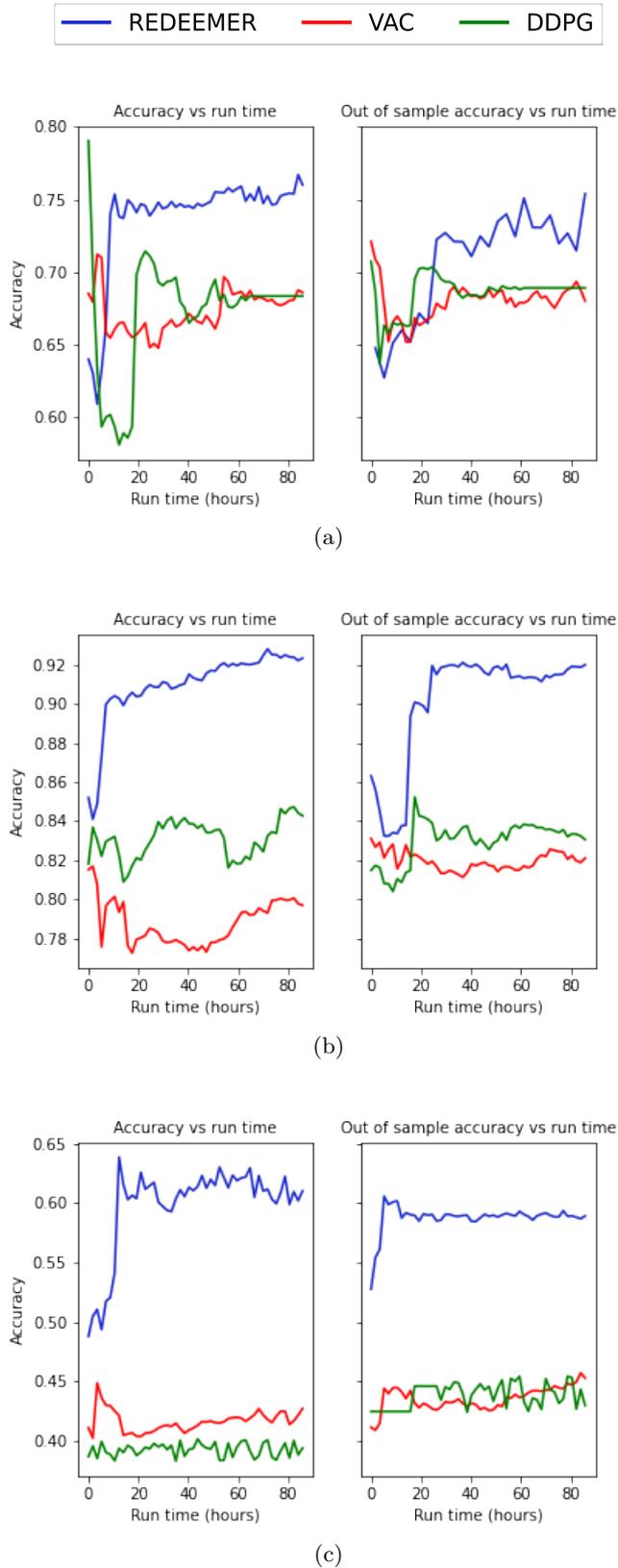

Figure 6.1: Quality of generated patterns: training accuracy and out of sample accuracy for pattern rank prediction during training. Results on the (a) StarPilot data-set, the (b) Football data-set, and the (c) GPU cluster data-set. Comparison of performance of the rank predictor module, REDEEMER outperforms alternatives in both metrics for all data-sets.

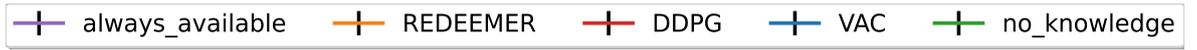

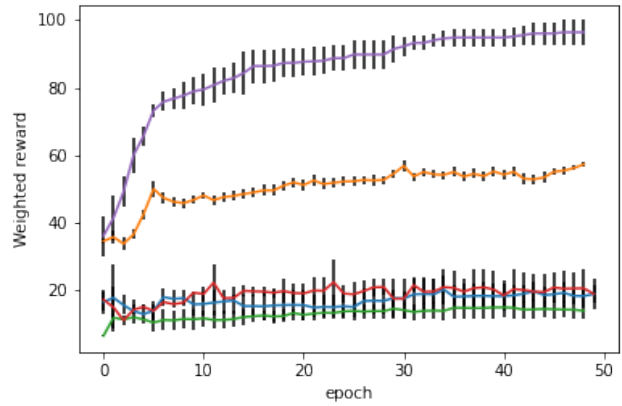

(a)

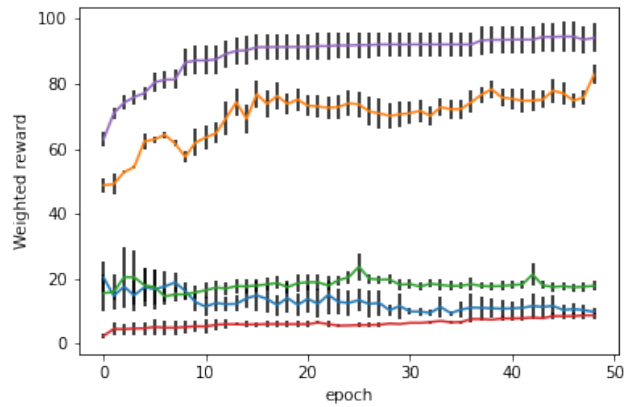

(b)

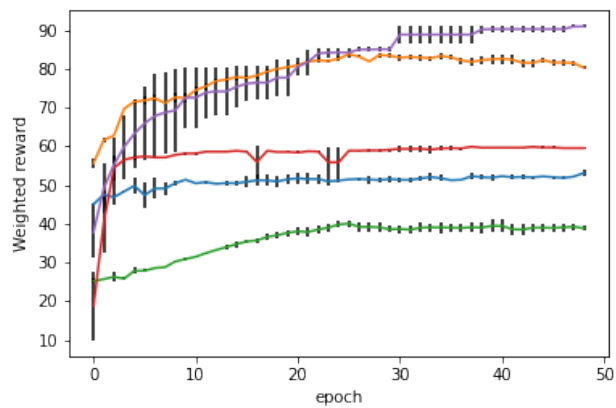

(c)

Figure 6.2: Quality of generated patterns: weighted model reward during training. Results on the (a) StarPilot data-set, the (b) Football data-set, and the (c) GPU cluster data-set. Comparison of the pattern constructor module of all models. Average weighted reward of patterns mined in the last epoch.

### 6.2.2 Availability Problem

Figure 6.3 and Table 6.6 show the comparison between experts with different availability. In this experiment, we compared three setups: 1) Expert is available in a limited capacity, 2) Expert is available in a limited capacity while also supplying some knowledge at initialization, 3) Expert is always available, and ranks every pattern generated during training. For setup 1, we used the **no_knowledge** architecture, for setup 2, we used both **REDEEMER** and **VAC**, and for setup 3, we used the **always_available** architecture.

In Figure 6.3, both the number of queries and the run-time metrics are shown for all datasets. Setups 1 and 3 can be thought of as the extremes of the run-time vs. number of queries trade-off. While both REDEEMER and VAC can find a middle ground, REDEEMER outperforms the baseline VAC method in both metrics.

We also compared the accuracy and out-of-sample accuracy of all four run-modes in Table 6.6. REDEEMER outperforms the two alternatives in both metrics, while the always_available cannot be surpassed since it is essentially the ground truth.

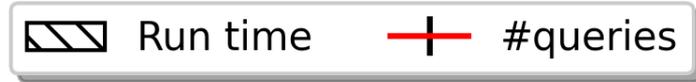

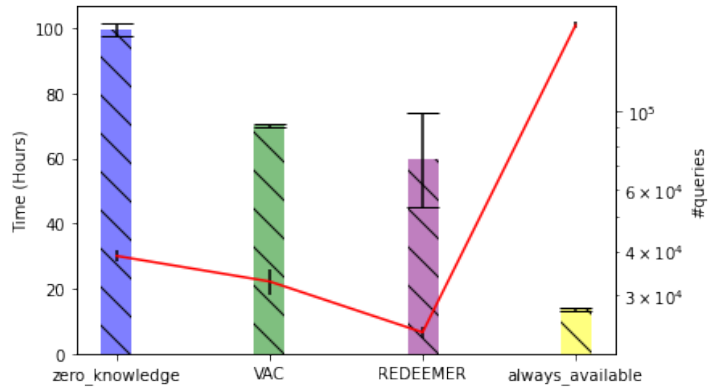

(a)

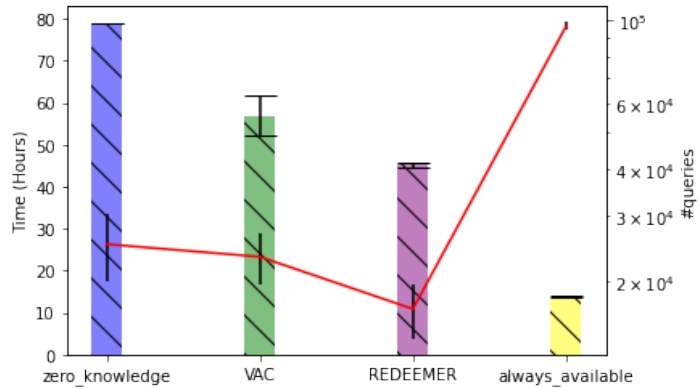

(b)

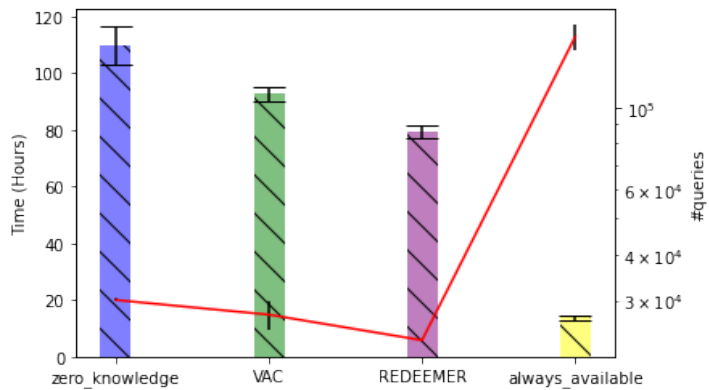

(c)

Figure 6.3: Availability problem: run-time and number of queries for expert trade-off. Results on the (a) StarPilot data-set, the (b) Football data-set, and the (c) GPU cluster data-set. Comparison of total run time and the number of queries needed for training when different run modes of REDEEMER are used. VAC used as a baseline for comparison. More run time is needed when no knowledge is supplied at initialization as the model converges slower. When an oracle replaces the prediction module the run time is shortest (since there is not a model to train) but the number of queries rises drastically.

| Run Mode | StarPilot | | Football | | GPU Cluster | |
|---|---|---|---|---|---|---|
| | Accuracy | Out of sample Accuracy | Accuracy | Out of sample Accuracy | Accuracy | Out of sample Accuracy |
| zero_knowledge | 0.395 | 0.319 | 0.802 | 0.752 | 0.221 | 0.324 |
| VAC | 0.685 | 0.681 | 0.821 | 0.791 | 0.427 | 0.453 |
| REDEEMER | 0.766 | 0.745 | 0.924 | 0.915 | 0.609 | 0.589 |
| always_available | 1.0 | 1.0 | 1.0 | 1.0 | 1.0 | 1.0 |

Table 6.6: Availability problem: Training accuracy and out of sample accuracy for pattern rank prediction. REDEEMER outperforms alternatives when it is initialized with simple patterns. Only the always_available run mode which uses an oracle for all rank prediction achieves better performance.

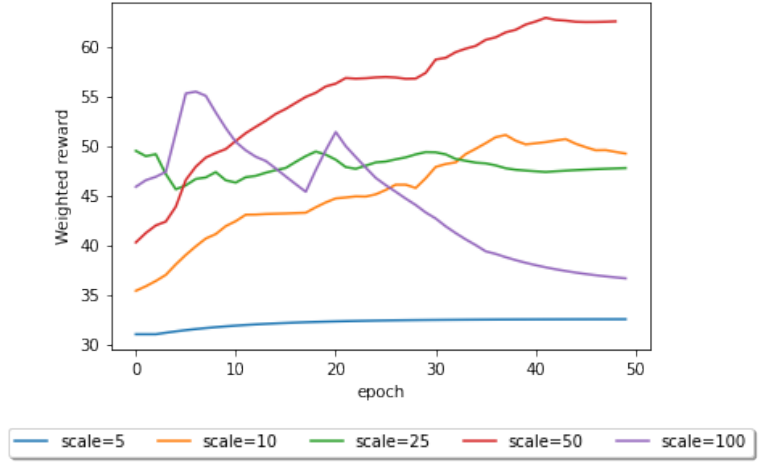

(a) Accuracy vs. scale size.

|           | Final Accuracy | Out of sample Accuracy |
|-----------|----------------|------------------------|
| scale=5   | 0.88           | 0.79                   |
| scale=10  | 0.812          | 0.676                  |
| scale=25  | 0.740          | 0.783                  |
| scale=50  | 0.766          | 0.745                  |
| scale=100 | 0.507          | 0.429                  |

(b) Reward vs. scale size.

Figure 6.4: Granularity problem: Ranking scale size comparison over the StarPilot data-set. The effects of the expert's expertise level on the mined pattern's reward trend and the final accuracy of pattern rank prediction. Low accuracy and reward achieved at scale=100 as not enough examples are gathered for each rank, and low reward for scale=5 as there is not enough distinguishability between patterns.

### 6.2.3 Granularity Problem

Although we generally use the number of queries sent for labeling as the metric for the amount of work required, the amount of work needed for a single query is impacted by how coarse or fine a label is expected to be (how specific the label should be).

In the following, we compare different runs of REDEEMER with different scale sizes for the pattern rankings. When scale size is small (coarse labels), the labeling is considered less challenging than in cases where scale size is large (fine labels) (e.g., $rank \in \{bad, ok, good\}$ is easier than $rank \in [1, 100]$). We compared runs where $scale\_size \in \{5, 10, 25, 50, 100\}$.

Figure 6.4a presents the weighted reward of REDEEMER during training, which was as explained in Section 6.2.1 and later scaled down by dividing each reward by the maximum scale size to normalize results and present them on the same axis. The final accuracy and out-of-sample accuracy are also presented in Table 6.4b.

Together the results from both figures paint a clear picture of the trade-offs for small and large pattern ranking scale sizes. Not only are large scale sizes harder for an expert to work with, but they also have lower accuracy since the classification task is over a larger number of classes. For example, when we set the scale size to 100, we received the lowest accuracy- 50.7%. On the other hand, we received high accuracy when using lower scale sizes (coarser labels). The scales of 5 and 10 were the only ones to achieve at least 80% in the prediction task.

Although the received accuracy was satisfactory, we found that when we used lower scale sizes, more patterns were grouped together with the same rating, causing less expressiveness in ranking and less distinguishability between them. This set a steeper learning curve that REDEEMER did not manage to climb. We observed that for scale=5, the weighted reward trend was almost flat, and no improvement could be seen. All figures refer to runs over the StarPilot data-set. The results over the other datasets were similar.

For other use cases, it is hard to suggest either a specific value or a formula for the scale size since it is dependent on both the expert's availability and expertise.

### 6.2.4 Reliability Problem

To further understand REDEEMER's performance in realistic use cases, we evaluated its accuracy metrics while it had access to experts with different levels of expertise.

To simulate imperfect experts, we sampled distortions in labeling from the normal distribution with fixed $\mu = 0$ and varying $\sigma$ values from $\{1, 5, 10, 20\}$. We also compared these runs with a scenario in which a perfect expert is available (labeled "no-noise" in the figures) and with **VAC**, which also utilizes a perfect expert.

Figures 6.5a and 6.5b present the final recorded in-sample and out-of-sample accuracy, respectively. It is clear that when the expertise level drops ($\sigma$ rises), RE-DEEMER achieves inferior results. Nevertheless, we can see that with some levels of noise ($\sigma \in [1, 5]$). REDEEMER still outperforms VAC, further highlighting the improvement achieved by the new architecture.

When low noise was applied to labels, there was a larger increase in out-of-sample performance compared to accuracy over the final $\mathcal{D}_i$. This fits with current-day methods for robust training using noisy labels [SKP+21, CC20], as less experienced experts give slightly incorrect rankings for patterns, which are indistinguishable from purposely noisy labels. Figures 6.5c and 6.5d present the same metrics in a dynamic graph with on-sample accuracy calculated over the ever-changing $\mathcal{D}_i$ group. The out-of-sample accuracy is calculated over the fixed $\mathcal{D}'$.

We can see that even when used with a moderately noisy expert, REDEEMER can improve during training and does well in generalizing results to out-of-sample data.

When used with $\sigma = 20$, we note that both trends do not represent a constant improvement, meaning the noise applied was more than REDEEMER could handle. For $\sigma > 20$, we found similar results, where the model did not show any clear improvement.

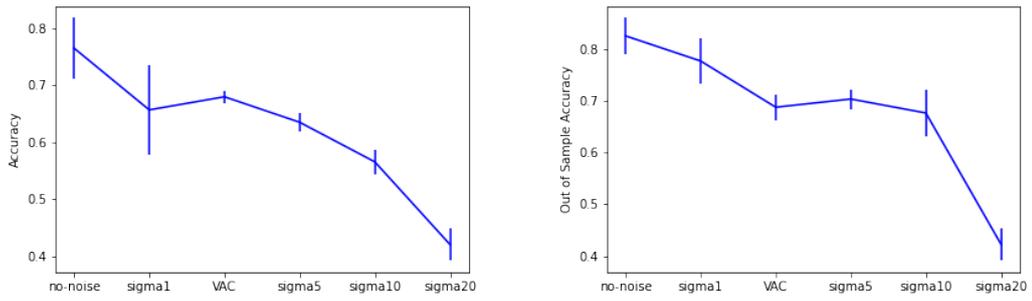

(a) Accuracy vs. expert noise final results.

(b) Out of Sample Accuracy vs. expert noise final results.

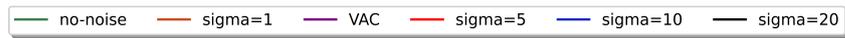

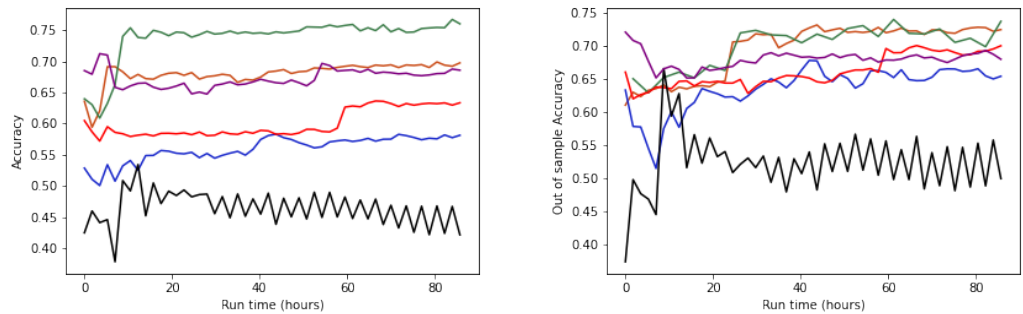

(c) Accuracy vs. expert noise during training.

(d) Out of Sample Accuracy vs. expert noise during training.

Figure 6.5: Reliability problem: accuracy vs. quality and reliability of the domain expert. The effects of the expert's errors in rating queries on the learning trend and the overall accuracy achieved.

# Chapter 7

# Conclusion

In this thesis, we formally defined the problem of pattern knowledge extension from a limited number of simple examples. We presented REDEEMER, which is the first work to solve this problem to the best of our knowledge.

From our thorough experiments, we believe the presented REDEEMER method is not only the first to solve the pattern knowledge expansion problem but also can be relevant for realistic use cases in which prior knowledge is minimal, expert availability is limited, and when expert's work is imperfect.

In summary, the results of our experimentation demonstrate the feasibility of the proposed REDEEMER approach.

**Future Work.** Future directions for our work include extending the *action-branching* architecture to allow the representation of larger multivariate action-spaces by using more complex data or allowing for more domain-specific operators to be used. We also plan to incorporate methods that combine information gathered from multiple sources that differ in availability and reliability.

# Bibliography

[ABNS06]    Asaf Adi, David Botzer, Gil Nechushtai, and Guy Sharon. Complex event processing for financial services. In *2006 IEEE Services Computing Workshops*, pages 7–12, 2006.

[ACBF02]    Peter Auer, Nicolò Cesa-Bianchi, and Paul Fischer. Finite-time analysis of the multiarmed bandit problem. *Machine Learning*, 47(2):235–256, 5 2002.

[AcT08]     Mert Akdere, Uundefinedur Çetintemel, and Nesime Tatbul. Plan-based complex event detection across distributed sources. *Proc. VLDB Endow.*, 1(1):66–77, aug 2008.

[AH14]      Charu C. Aggarwal and Jiawei Han. *Frequent Pattern Mining*. Springer Publishing Company, Incorporated, 2014.

[AIS93]     Rakesh Agrawal, Tomasz Imielinski, and Arun Swami. *Mining Association Rules Between Sets of Items in Large Databases, SIGMOD Conference*, volume 22, page 207. Washington D.C. USA, 06 1993.

[Aue03]     Peter Auer. Using confidence bounds for exploitation-exploration trade-offs. *J. Mach. Learn. Res.*, 3(null):397–422, March 2003.

[BDO19]     Ralf Bruns, Jürgen Dunkel, and Norman Offel. Learning of complex event processing rules with genetic programming. *Expert Systems with Applications*, 129:186–199, 2019.

[BEE+10]    Marion Blount, Maria R. Ebling, J. Mikael. Eklund, Andrew G. James, Carolyn McGregor, Nathan Percival, Kathleen Smith, and Daby Sow. Real-time analysis for intensive care: Development and deployment of the artemis analytic system. *IEEE Engineering in Medicine and Biology Magazine*, 29(2):110–118, March 2010.

[Bie20]     Lukas Biewald. Experiment tracking with weights and biases, 2020. Software available from wandb.com.

[BOSB10]   Kay Henning Brodersen, Cheng Soon Ong, Klaas Enno Stephan, and
           Joachim M. Buhmann. The balanced accuracy and its posterior distri-
           bution. In *2010 20th International Conference on Pattern Recognition*,
           pages 3121–3124, 2010.

[CC20]     Filipe R. Cordeiro and Gustavo Carneiro. A survey on deep learning
           with noisy labels: How to train your model when you cannot trust on
           the annotations?, 2020.

[CHHS20]   Karl Cobbe, Chris Hesse, Jacob Hilton, and John Schulman. Leveraging
           procedural generation to benchmark reinforcement learning. In *Interna-
           tional conference on machine learning*, pages 2048–2056. PMLR, 2020.

[DAEvH+16] Gabriel Dulac-Arnold, Richard Evans, Hado van Hasselt, Peter Sunehag,
           Timothy Lillicrap, Jonathan Hunt, Timothy Mann, Theophane Weber,
           Thomas Degris, and Ben Coppin. Deep reinforcement learning in large
           discrete action spaces, 2016.

[DP18]     Melanie Ducoffe and Frederic Precioso. Adversarial active learning for
           deep networks: a margin based approach, 2018.

[FVLR+17]  Philippe Fournier Viger, Chun-Wei Lin, Uday Rage, Yun Sing Koh, and
           Rincy Thomas. A survey of sequential pattern mining. *Data Science and
           Pattern Recognition*, 1:54–77, 02 2017.

[FVLV+17]  Philippe Fournier-Viger, Jerry Chun-Wei Lin, Bay Vo, Tin Truong Chi,
           Ji Zhang, and Hoai Bac Le. A survey of itemset mining. *WIREs Data
           Mining and Knowledge Discovery*, 7(4):e1207, 2017.

[GA19]     László Gadár and János Abonyi. Frequent pattern mining in multidi-
           mensional organizational networks. *Scientific Reports*, 9:3322, 03 2019.

[GPAM+14]  Ian J. Goodfellow, Jean Pouget-Abadie, Mehdi Mirza, Bing Xu, David
           Warde-Farley, Sherjil Ozair, Aaron Courville, and Yoshua Bengio. Gen-
           erative adversarial networks, 2014.

[HP00]     Jiawei Han and Jian Pei. Mining frequent patterns by pattern-growth:
           Methodology and implications. *SIGKDD Explorations*, 2:14–20, 01 2000.

[HRE+18]   Ashley Hill, Antonin Raffin, Maximilian Ernestus, Adam Gleave, Anssi
           Kanervisto, Rene Traore, Prafulla Dhariwal, Christopher Hesse, Oleg
           Klimov, Alex Nichol, Matthias Plappert, Alec Radford, John Schulman,
           Szymon Sidor, and Yuhuai Wu. Stable baselines. `https://github.com/`
           `hill-a/stable-baselines`, 2018.

[Kol21]    Ilya Kolchinsky. OpenCEP: open-source library and framework providing
           advanced complex event processing (CEP) capabilities., 2021.

[KS18a]     Ilya Kolchinsky and Assaf Schuster. Efficient adaptive detection of complex event patterns, 2018.

[KS18b]     Ilya Kolchinsky and Assaf Schuster. Join query optimization techniques for complex event processing applications, 2018.

[LHP+19]    Timothy P. Lillicrap, Jonathan J. Hunt, Alexander Pritzel, Nicolas Heess, Tom Erez, Yuval Tassa, David Silver, and Daan Wierstra. Continuous control with deep reinforcement learning, 2019.

[LQSC12]    Ninghui Li, Wahbeh Qardaji, Dong Su, and Jianneng Cao. Privbasis: Frequent itemset mining with differential privacy. *Proceedings of the VLDB Endowment*, 5, 07 2012.

[MBM+16]    Volodymyr Mnih, Adrià Puigdomènech Badia, Mehdi Mirza, Alex Graves, Timothy P. Lillicrap, Tim Harley, David Silver, and Koray Kavukcuoglu. Asynchronous methods for deep reinforcement learning, 2016.

[MKE+15]    Nijat Mehdiyev, Julian Krumeich, David Enke, Dirk Werth, and Peter Loos. Determination of rule patterns in complex event processing using machine learning techniques. *Procedia Computer Science*, 61:395–401, 12 2015.

[MM09]      Yuan Mei and Samuel Madden. Zstream: A cost-based query processor for adaptively detecting composite events. In *Proceedings of the 2009 ACM SIGMOD International Conference on Management of Data*, SIGMOD '09, page 193–206, New York, NY, USA, 2009. Association for Computing Machinery.

[MSIB20]    Nina Mazyavkina, Sergey Sviridov, Sergei Ivanov, and Evgeny Burnaev. Reinforcement learning for combinatorial optimization: A survey, 2020.

[MTZ17]     Raef Mousheimish, Yehia Taher, and Karine Zeitouni. Automatic learning of predictive cep rules: Bridging the gap between data mining and complex event processing. In *Proceedings of the 11th ACM International Conference on Distributed and Event-Based Systems*, DEBS '17, page 158–169, New York, NY, USA, 2017. Association for Computing Machinery.

[MZJ13]     Christopher Mutschler, Holger Ziekow, and Zbigniew Jerzak. The debs 2013 grand challenge. In *Proceedings of the 7th ACM International Conference on Distributed and Event-Based Systems*, DEBS '13, page 289–294, New York, NY, USA, 2013. Association for Computing Machinery.

[Nog20]      Fernando Nogueira. Bayesian Optimization: Open source constrained global optimization tool for Python, 2020.

[PVS11]      Adrian Paschke, Paul Vincent, and Florian Springer. Standards for complex event processing and reaction rules. volume 7018, pages 128–139, 01 2011.

[RXC+21]     Pengzhen Ren, Yun Xiao, Xiaojun Chang, Po-Yao Huang, Zhihui Li, Brij B. Gupta, Xiaojiang Chen, and Xin Wang. A survey of deep active learning, 2021.

[SB18]       Richard S. Sutton and Andrew G. Barto. *Reinforcement Learning: An Introduction*. The MIT Press, second edition, 2018.

[Set09]      Burr Settles. Active learning literature survey. Computer Sciences Technical Report 1648, University of Wisconsin–Madison, 2009.

[SKP+21]     Hwanjun Song, Minseok Kim, Dongmin Park, Yooju Shin, and Jae-Gil Lee. Learning from noisy labels with deep neural networks: A survey, 2021.

[SZGW20]     Changjian Shui, Fan Zhou, Christian Gagné, and Boyu Wang. Deep active learning: Unified and principled method for query and training. In Silvia Chiappa and Roberto Calandra, editors, *Proceedings of the Twenty Third International Conference on Artificial Intelligence and Statistics*, volume 108 of *Proceedings of Machine Learning Research*, pages 1308–1318. PMLR, 26–28 Aug 2020.

[TPK19]      Arash Tavakoli, Fabio Pardo, and Petar Kormushev. Action branching architectures for deep reinforcement learning, 2019.

[WDR06]      Eugene Wu, Yanlei Diao, and Shariq Rizvi. High-performance complex event processing over streams. In *Proceedings of the 2006 ACM SIGMOD International Conference on Management of Data*, SIGMOD '06, page 407–418, New York, NY, USA, 2006. Association for Computing Machinery.

[Wen18]      Lilian Weng. Policy gradient algorithms. *lilianweng.github.io/lil-log*, 2018.

[WLC+20]     Xinyue Wang, Bo Liu, Siyu Cao, Liping Jing, and Jian Yu. Important sampling based active learning for imbalance classification. *Science China Information Sciences*, 63(8):182104, 7 2020.

[WvAHS09]    Alexander Widder, Rainer von Ammon, Gerit Hagemann, and Dirk Schönfeld. An approach for automatic fraud detection in the insurance domain. In *Intelligent Event Processing, Papers from the 2009*

*AAAI Spring Symposium, Technical Report SS-09-05, Stanford, California, USA, March 23-25, 2009*, pages 98–100. AAAI, 2009.

[YC05]    Chung-Ching Yu and Yen-Liang Chen. Mining sequential patterns from multidimensional sequence data. *IEEE Transactions on Knowledge and Data Engineering*, 17(1):136–140, 2005.

[YK09]    Lexiang Ye and Eamonn Keogh. Time series shapelets: A new primitive for data mining. In *Proceedings of the 15th ACM SIGKDD International Conference on Knowledge Discovery and Data Mining*, KDD '09, page 947–956, New York, NY, USA, 2009. Association for Computing Machinery.

[Zak00]   M.J. Zaki. Scalable algorithms for association mining. *IEEE Transactions on Knowledge and Data Engineering*, 12(3):372–390, 2000.

[ZCMR21]  Huayi Zhang, Lei Cao, Samuel Madden, and Elke Rundensteiner. Lancet: Labeling complex data at scale. *Proc. VLDB Endow.*, 14(11):2154–2166, jul 2021.

ניסויים נוספים מציגים את יכולת המערכת להתמודד עם בעיות שנפוצות בסיטואציות שאינן אקדמיות, כגון זמינות מוגבלת של מומחה ביחס למצופה, רמת יכולת נמוכה של מומחה במענה על השאילתות של המודל, חוסר יכולת של מומחה להבדיל בין ערכן של תבניות בצורה מדוייקת (כלומר יכולת הבדלה גסה וכללית מהנדרש). מטרת קבוצת ניסויים זו היא להציג את הרלוונטיות של מערכת REDEEMER אל סיטואציות ריאליות שקיימות בתחומים בהן כיום אין מספיק ידע קיים להשתמש במערכות עיבוד אירועים מורכבים, שכן מטרת המחקר שלנו הייתה למצוא פיתרון אפקטיבי ככל הניתן שיאפשר את קידום העולם של CEP ופישוטו הקושי בשילובו במערכות קיימות.

בתיזה זו, אנו מציג גישה אלטרנטיבית וחדשה שמניחה ידע מוגבל אחר שקיים על ידי המומחה, ומנסה לגשר על הפער של התבניות הלא ידועות בצורה אחרת.  בעוד שבשיטות הקיימות שהצגנו ישנה הנחה שהמומחה יודע במידה מסוימה (יודע לזהות בזמן אמת מופעים וכן להביא מבנה מקורב של התבנית) על התבנית אותה הוא מחפש, אנחנו נתמודד עם סיטואציה שידע כזה לא קיים, אך כן קיים ידע (בדמות תבניות מפורשות) על תבניות פשוטות ופחות מעניינות בשטף הנתונים. בעוד השיטות הקיימות הצליחו לחדד ולעדן את התבנית שכבר הייתה ידועה למומחה, אנו נרצה למצוא תבניות חדשות לגמרי, שלא היו ידועות כלל למומחה לפני התחלת התהליך, ובכך נאפשר מציאה של תבניות חשובות, נפוצות ומעניינות חדשות שאותן אפשר לנצל בכדי להפיק מידע רב על המערכת.

הגישה שלנו מסתמכת על שתי שיטות המוכלות תחת הקטגוריה של למידה עמוקה, ונקראות למידה מחיזוקים (reinforcement learning) ולמידה אקטיבית (active learning), שתי שיטות פופולאריות מאוד הזוכות להכרה ספרותית רבה ושקיים בהן מחקר נרחב ומעמיק בשנים האחרונות. למידה מחיזוקים היא שיטת למידה מונחית בצורה חלקית, ומטרה ללמוד סוכן מדיניות עבודה כלשהי שמובילה לתוצאות אידיאליות בהינתן קלט מסוים.  למידה אקטיבית היא עולם מיוחד בלמידה עמוקה המתעמסק בסיטואציות בהן המידע הדרוש לאימון מודל אינו ידוע מראש, אלא מתקבל בצורה הדרגתית ותוך כדי הליך הלמידה על ידי אינטראקציה עם גורם מומחה שמוסיף ומתייג דוגמאות נבחרות אותן בחר המודל לתשאל אותו לגביהן.  מערכות המשתמשות בלמידה אקטיבית מנסות למצוא לא רק איך ללמוד לבצע משימה מסוימה אותה המודל צריך לבצע, אלא גם איך לשאול שאלות, כלומר להבין את הדרך הטובה ביותר לברור שאילות למומחה, כדי להפיק תועלת גדולה ככל הניתן מכל תשובה שלו, ובכך להקטין את חוסר הוודאות בפרדיקציה של המודל על כל קלט.

בתיזה זו, אנו נציג את מערכת להסקת תבניות נפוצות ומשמעותיות חדשניות לאפשר אינטגרציה של מערכות CEP באפליקציות חדשות.  המערכת מכונה בשם REDEEMER, קיצור של REinforcement baseD cEp pattErn MinER. המערכת שלנו מקבלת כקלט כללם מופעים בשטף המידע של מספר קטן מאוד של תבניות ידועות מראש שאותן זיהה המומחה, וכדירוג של כמה מעניינות הדוגמאות עבורו. תבניות אלו אמורות להיות פשוטות, וקלות לזיהוי, ואינן התבניות אותן המומחה שואף לנצל במערכת CEP. לאחר מכן, בעזרת מודל מבוסס למידה מחיזוקים, אנו מאמנים סוכן לבנות תבניות מורכבות חדשות, אשר צריכות להיות לא רק מעניינות עבור המומחה, אלא גם רלוונטיות מספיק ובעלות נוכחות מספקת בשטף המידע (בכדי שישימוש בהן יניץ תועלת גדולה ככל הניתן). המערכת שלנו כוללת גם רכיב מבוסס למידה אקטיבית לשערוך ערכה של כל תבנית, רכיב זה משתמש בדוגמאות שניתנו על ידי המומחה בכדי לשערך כל תבנית, וכן יוכל להעביר לאורך תהליך האימון של המערכת קובצת שאילתות נוספות אל המומחה בכדי להגדיל את קבוצת הדוגמאות שנמצאת בידו ובכך לקבל תמונה ברורה יותר של מרחב התבניות ולספק שערוך טוב יותר.  בסוף תהליך האימון, המערכת מחזירה למומחה את קבוצת התבניות הטובות ביותר שמצאה מבחינת עניין (התאמה לצרכים של המומחה, כפי ששוערכה בעזרת הדוגמאות שסיפק) וכן מבחינת רלוונטיות (כמה נפוצות היו התבניות בשטף המידע, ככל שנפוצות יותר, שימוש בהן יפיק תועלת גבוהה יותר), התבניות מוחזרות בשפה דמוית SQL הניתנת לקריאה והבנה עבור מומחה תוכן ששולט בעולם הבעיה.

בפרק הניסויים המוצג בתיזה זו, מתוארים מגוון רחב של ניסויים שביצענו על גבי שלושה דאטהסטים פומבייסומעוגנים. הניסויים שלנו נועדו לבדוק את איכות התבניות שמערכת REDEEMER מסוגלת לגלות במהלך הליך אימון ארוך, וכן את הדיוק שלה בשערוך התועלת של תבניות חדשות שכרתה.

# תקציר

גילוי תבניות מורכבות בזרמי מידע מאסיביים יכול לשמש בתחומים/יישוגילוי תבניות מורכבות בזרמי מידע מאסיביים יכול לשמש בתחומים ואפליקציות רבות על מנת לבצע זיהוי יעיל של תופעות חשובות ויצירת התרעות קריטיות על איומים פוטנציאיליים, כשלים שעומדים להתרחש או הזדמניות חשובות שניתן לנצל. טכנולוגית לעיבוד אירועים מורכבים (הידועה גם בשם מערכות CEP) נמצאות בשימוש נרחב, מאפשרות זיהוי יעיל בזמן אמת של תבניות מורכבות המופיעות בזרמי מידע מאסיביים המגיעים ממקורות מידע רבים בקצב מהיר ולא צפוי. כבר היום מערכות שכאלו מנוצלות בתחומים רבים ומאפשרות שיפור ביצוע ניכר. בין היתר מערכות לעיבוד אירועים מורכבים קיימות קיימות היום במערכות פיננסיות, בשרותי אבטחה ברשת, במערכות גילוי הונאה, מערכות המנטרות תעבורה, מערכות ניטור מידע רפואי ועוד מערכות רבות אחרות.

למרות התועלת הגדולה שניתן לקבל בעזרת שימוש ממערכות לעיבוד אירועים מורכבים, קיים קושי גדול לשלב אותן באפליקציות רבות. כיום, כל הפתרונות הקיימים מניחים שהתבניות המעניינות שבהן אפשר להשתמש ידועות מראש ומוזנות עבור המערכת באתחול. אולם ידע שכזה לא קיים במרבית הסיטואציות, שכן תבניות מעניינות לרוב יהיו מורכבות מאוד ויכללו תנאים רבים בין אירועים שונים ועל כן יהיו קשות לזיהוי על ידי בני אדם. בנוסף, בניית כל תבנית מעניינת שכזאת היא משימה קשה מאוד, וגם במידה שהיא אפשרית בסיטואציה נתונה, היא תדרוש שעות עבודה רבות וידע נרחב בעולם הבעיה, אין מספיק מומחים שיכולים לעמוד בדרישה לעבודות המתבצעות כיום, וכן כל שעת עבודה של מומחה היא יקרה מאוד.

בשנים האחרונות התחילה העבודה שנועדה לגשר על הפער הזה, מתוך מטרה לאפשר אינטגרציה של מערכות עיבוד אירועים מורכבים במגוון אפליקציות חדשות. הפתרונות הקיימים בררו בגישה שבה המומחה הנדרש להגדרת התבנית זמן רב אך הידע שלו מוגבל במידה מסוימת. סוג אחד של פתרונות מניח שבשטף המידע ישנה רק תבנית מעניינת אחת אותה המומחה מנסה למצוא, וכן למרות שאינו יודע להגדיר את התבנית בצורה מפורשת הם מניחים שהוא מסוגל למצוא מופעו שלה ולסמנם עבור מערכות לומדות או עבור אלגוריתמים גנטיים שמנסים להסיק ולמצוא את התבנית המדויקת ביותר שאותה המומחה מנסה לגלות.

טכניקה אחרת שבה נקטו עבודות אחרות מניחות שלמומחה יש מבנה מקורב של מספר תבניות אותן הוא מגדיר כמעניינות, וכן מניחות שלמומחה תהיה היכולת לזהות בזמן אמת מופעים של כל אחת מהתבניות, וכן לסווג לאיזה תבנית שייך המופע שהוא זיהה. בעבודות הללו המערכות לרוב משתמשות באלגוריתמי למידה עמוקה שמטרתם לחדד ולעדן את המבנה המקורב לתבניות ידוע למומחה. המטרה הסופית של מערכות אלו היא להגיע לא לתבנית אחת, אלא למצוא את סט התבניות המדויקות ביותר, שיוכלו לתרום למערכת בצורה נרחבת ולאפשר את מציאת התבניות הטובות ביותר.

# כריית תבניות נפוצות במערכות CEP

חיבור על מחקר

**גיא שפירא**

# כריית תבניות נפוצות במערכות CEP

גיא שפירא